\documentclass[preprint,12pt]{elsarticle}
\usepackage[pdfborder={0 0 0}]{hyperref}
\usepackage{bm}
\usepackage{placeins}
\usepackage[title]{appendix}
\usepackage{graphics,fancyhdr,graphicx,subfigure}
\usepackage{subfigure}
\usepackage{amsthm,amsmath,amsfonts,latexsym,amssymb}
\usepackage{lineno}
\usepackage[ruled,linesnumbered]{algorithm2e}
\usepackage[margin=1in]{geometry}
\usepackage{multirow,booktabs}
\usepackage{listings}
\usepackage[table]{xcolor}
\usepackage{lscape}
\usepackage{tabularx}
\biboptions{numbers,sort&compress}
\usepackage{comment}
\DeclareMathOperator*{\argmin}{\arg\!\min}

\def\equationautorefname~#1\null{%
  Eq.~(#1)\null
  }
\def\subfigureautorefname~#1\null{%
  Fig.~#1\null
}

\definecolor{listinggray}{gray}{0.9}
\definecolor{lbcolor}{rgb}{0.9,0.9,0.9}
\definecolor{Darkgreen}{RGB}{0,100,0}

\def \bb{\bm{\beta}}

\def \fb{\bm{f}}

\def \ub{\mathbf{u}}

\def \wb{\bm{w}}

\def \xb{\bm{x}}

\def \yb{\bm{y}}

\DeclareMathOperator{\real}{\mathbb{R}}

\begin{document}
\abovedisplayskip=6.0pt
\belowdisplayskip=6.0pt
\begin{frontmatter}

\title{A physics-informed variational DeepONet for predicting the crack path in brittle materials}


\address[gk]{Division of Applied Mathematics, Brown University, Providence, RI}
\address[my]{Center for Biomedical Engineering, Brown University, Providence, RI}
\address[my2]{School of Engineering, Brown University, Providence, RI}
\address[yy]{Department of Mathematics, Lehigh University, Bethlehem, PA}

\author[gk]{Somdatta Goswami}\ead{somdatta\_goswami@brown.edu}
\author[my,my2]{Minglang Yin}\ead{minglang\_yin@brown.edu}
\author[yy]{Yue Yu}\ead{yuy214@lehigh.edu}
\author[gk,my2]{George Em Karniadakis\corref{cor1}}\ead{george\_karniadakis@brown.edu}
\cortext[cor1]{Corresponding author.}

\begin{abstract}
\noindent
Failure trajectories, identifying the probable failure zones, and damage statistics are some of the key quantities of relevance in brittle fracture applications. High-fidelity numerical solvers that reliably estimate these relevant quantities exist but they are computationally demanding requiring a high resolution of the crack. Moreover, independent intensive simulations need to be carried out even for a small change in domain parameters and/or material properties. Therefore, fast and generalizable surrogate models are needed to alleviate the computational burden but the discontinuous nature of fracture mechanics presents a major challenge to developing such models. We propose a physics-informed variational formulation of DeepONet (V-DeepONet) for brittle fracture analysis. V-DeepONet is trained to map the initial configuration of the defect to the relevant fields of interests (e.g., damage and displacement fields). Once the network is trained, the entire global solution can be rapidly obtained for any initial crack configuration and loading steps on that domain. While the original DeepONet is solely data-driven, we take a different path to train the V-DeepONet by imposing the governing equations in variational form and we also use some labelled data. We demonstrate the effectiveness of V-DeepOnet through two benchmarks of brittle fracture, and we verify its accuracy using results from high-fidelity solvers. Encoding the physical laws and also some data to train the network renders the surrogate model capable of accurately performing both interpolation and extrapolation tasks, considering that fracture modeling is very sensitive to fluctuations. The proposed hybrid training of V-DeepONet is superior to state-of-the-art methods and can be applied to a wide array of dynamical systems with complex responses.
 
\end{abstract}
\end{frontmatter}

\section{Introduction}
\label{sec:introduction}

Damage evolution in realistic structures is a complex phenomenon. Accurately representing this behaviour relies on complex and computationally expensive high-fidelity models. Traditional approaches in computational science have undergone remarkable growth and progress but they still operate under stringent requirements. More often than not they require precise knowledge of an underlying model that describes conservation. Moreover, most existing numerical methods utilize spatial discretization and thus are prone to the curse of dimensionality. Newer alternatives such as phase-field \cite{kuhn2010continuum,borden2012phase,nguyen2015phase,borden2014higher} and peridynamics \cite{emmrich2007well,silling2000reformulation,yu2018partitioned,haghighat2021nonlocal,trask2019asymptotically,yu2021asymptotically} models have been able to predict the outcome of carefully controlled experiments; however, the complexity of those models comes at a substantially higher computational cost. Furthermore, operational conditions and material  properties in the field can have a significant deviation from those in a controlled laboratory environment, which impairs the reliability of the computed results. In order to assess the likelihood of occurrence of cracks and their possible effects for a range of possible parameters and operating conditions, a large number of high-fidelity simulations is required, which are typically computationally prohibitive. 

Surrogate models \cite{psichogios1992hybrid,haghighat2021physics,peherstorfer2017combining,hou2018novel,van2021neural, martinez2021machine,you2021data,You2021,you2021MD} have received a lot of attention because of their ability to quantitatively capture the fundamental attributes of high-fidelity models while significantly improving the computational efficiency. The main challenge associated with constructing a surrogate model for fracture analysis comes from the inherent discontinuous nature of the physical phenomena. Conventionally, surrogate models require smoothness in the system response, while models for fracture mechanics are very sensitive to fluctuations in model properties and hence exhibit irregular behaviour. This problem has prompted a burgeoning literature on reduced-order approaches \cite{beran2001reduced, amsallem2008interpolation, amsallem2009method}, which use existing data sets to create fast emulators often at the expense of accuracy, stability, and generalization. The goal of the present work is to develop a flexible generalized prognosis framework for high-fidelity crack growth models. To this end, we will use deep neural networks (DNNs)to infer the generalized solution of the governing partial differential equations (PDEs) that describe the fracture mechanisms.

Deep learning allows overparametrized neural networks with several processing layers to learn multiple levels of abstraction for representations of the raw input data. These networks are known to be particularly good at supervised learning tasks, which typically necessitate the availability of huge volumes of labelled data. However, data collecting is generally prohibitively expensive in many engineering applications, and the amount of available data is typically minimal. As a result, in this “sparse data” environment, it is critical to use domain knowledge to reduce the demand for labelled training data, or even to train deep learning models using only constraints rather than data. Physics-informed neural networks (PINNs) \cite{raissi2019physics,samaniego2020energy,karniadakis2021physics,cai2021physics} use these constraints to encapsulate the output's specific structure and qualities, which are known to hold due to domain knowledge, such as known physical laws like conservation of momentum, mass, and energy. This approach takes advantage of the expressivity of DNNs to approximate any continuous function. To efficiently approximate the solution of PDEs with discontinuity, variational energy-based PINN (VE-PINN) was proposed in \cite{goswami2020transfer,goswami2020adaptive_dem}, where the network is trained by minimizing the variational energy of the system (defined using the weak formulation). VE-PINN opened a new paradigm for solving fracture problems with modern neural network architectures, which may be a promising alternative to traditional numerical methods, such as finite-difference and finite-volume methods as it reduced the computational burden of dense discretization. Despite the promise and collection of impressive results for accurately estimating the crack path in brittle fracture problems using a sparse discretization, VE-PINN bears a formidable cost as independent simulations need to be performed for every different domain geometry, input parameters, or initial/boundary conditions (I/BCs). The bottleneck of VE-PINN is quite similar to traditional numerical methods, hence the use of VE-PINN as a surrogate model for approximating the crack path for different initial conditions or for comprehensive uncertainty quantification is practically not feasible.

As the ML revolution continues to sweep the scientific world, a new wave of strategies for expediting the simulation of PDEs \cite{yin2021non,jagtap2020extended,zhang2020physics,weinan2018deep,bar2019unsupervised,guo2016convolutional,zhu2018bayesian,adler2017solving,bhatnagar2019prediction,li2020multipole} is being offered. Discovering PDEs just from data without any prior information is difficult in a generic situation. In most practical circumstances, it is necessary to have a surrogate model of the PDE solution operator that can simulate PDE solutions repeatedly for varied I/BCs, rather than discovering the PDE in an explicit form, to handle this difficulty. DeepONet proposed in \cite{lu2021learning} is one of the possible ways to learn the PDE solution operators from the labelled input-output datasets. The idea of DeepONet is motivated by the universal approximation theorem for operators. This defines a new and relatively under-explored realm for DNN-based approaches that map infinite-dimensional functional spaces rather than finite-dimensional vector spaces (functional regression). The computational model consists of two DNNs, one encodes the input function at fixed sensor points (branch net), while another for the location of the output function (trunk net). In this work, we propose a variational energy-based framework of DeepONet to parametrize and learn the solution operator that maps multiple initial conditions of the crack in the domain as input function to the branch net to their associated solutions at the locations embedded in the trunk net, thus overcoming the fundamental challenge of VE-PINN. The proposed deep learning model is trained by minimizing a hybrid loss function constructed using the PDEs defining the variational formulation of phase field approach, its associated I/BCs and relatively small input-output datasets generated using the in-house high-fidelity solvers. Once the model has been trained on a set of specific conditions (initial configurations, loadings steps, etc.\ ), it can be used to build the global PDE solution to predict quantities of interest using a simple iterative approach in which the previous time-step's prediction is utilized as an initial condition for the current time step. We demonstrate that this approach can effectively enable the integration of evolution equations subject to a range of multiple initial conditions with good generalization accuracy, all at a fraction of the computational cost needed by classical numerical solvers. The main advantage of our approach is that we make no assumptions on the regularity of the full model and as we demonstrate herein the ability to generalize and predict outcomes for inputs outside the distribution (extrapolation),

We demonstrate the performance of the surrogate model on two benchmark problems of fracture: crack growth under tensile loading (Mode-I) and shear loading (Mode-II). In this work, we have used the phase field approach to model fracture within the framework of isogeometric analysis (IGA) developed in \cite{goswami2020adaptive} to generate data. However, the method presented here is not restricted to any specific fracture model or mode of failure or data generated using specific high-fidelity solvers.  Even though the surrogate model proposed in this work is applicable to various evolution equations, e.g., time dependent problems and fracture dynamics, the integration of variational formulation makes its a perfect choice for PDEs with discontinuous nature which manifests in both spatial domain and system response for different material properties. The remainder of the paper is organized as follows. In \autoref{sec:concept_fracture}, we discuss the problem statement for phase-field modeling of brittle fracture using the variational energy formulation. In \autoref{sec:deeponet}, we provide an overview of the DeepONet framework put forth in \cite{lu2021learning}. Implementation of the variational formulation within DeepONet and the concept of hybrid loss function is discussed in \autoref{sec:vedeeponet}. The details of the construction of the proposed surrogate model within V-DeepONet, its implementation and its numerous application are elaborated in \autoref{sec:surrogate_fracture}, with data generation procedures described in \autoref{sec:data_generation}. In \autoref{sec:numericals} different aspects of the surrogate model illustrating its performance are tested through numerical problems. Each numerical example in the manuscript is accompanied with a detailed discussion about the neural network architecture we employed as well as details about its training process. Finally, \autoref{sec:conclusion} presents the concluding remarks and possibilities future work.
In the Appendices we present more details and additional cases for the interested readers to be able to reproduce our results. 

\section{Phase field modeling of fracture}
\label{sec:concept_fracture}

In recent times, phase field modeling approaches \cite{jacqmin1999calculation} have been extensively used in science and engineering to model a variety of phenomena. Modeling fracture, using the phase field approach, involves the integration of two fields, namely the vector-valued elastic field and the scalar-valued phase field. While crack nucleation may depend on stress, the propagation of cracks requires an increase in the fracture energy or the surface energy, $\Psi_c$ of a solid \cite{A.Griffith1921}. Hence, the energy criteria are used in the study of fracture using the phase field approach \cite{Francfort1998}. The physical domain, $\Omega \subset \real^d$, is defined with the external boundary, $\partial \Omega \subset \real^{d-1}$, where $d$ denotes the number of spatial dimensions, $d \in \{1,2,3\}$. In a quasi-static loading regime, the total energy functional, $\mathcal{E}$, can be written as \cite{wu2018phase}:
\begin{equation}\label{eq:totalenergy}
    \begin{split}
        \mathcal{E} &= \Psi_e + \Psi_c -\mathcal{P}_{ext},\\
\text{where}\;\;\Psi_e &=  \int_{\Omega}{\psi_e\left(\bm{\epsilon}\left(\bm w\right),\phi\left(\bm{x}\right)\right)} \;d\Omega,\\
       \Psi_c &= \int_{\Omega}{G_c\Theta\left(\phi(\bm{x}), l_0\right)} d\Omega\;\;\approx \int_{\Gamma_d}{G_c}\; d\Gamma,\\
\text{and}\;\;\mathcal{P}_{ext} &= \int_{\Omega}{\bm{f}\cdot \bm{w}}\; d\Omega + \int_{\partial\Omega_N}{\bm{t_N}\cdot \bm{w}}\;d\Gamma.
    \end{split}    
\end{equation}

In \autoref{eq:totalenergy}, $\Psi_e$ is the stored elastic strain energy, $d\Gamma$ denotes the integral on the co-dimensional one space (curves for $d=2$ an surfaces for $d=3$), $\Gamma_d$ is the evolving internal discontinuity boundary and $\psi_e$ is the strain energy density functional expressed in terms of the linearized strain tensor, $\bm{\epsilon}(\bm w)$, where $\bm{w}$ denotes the displacement field and a continuous scalar parameter $\phi(\bm{x})$ denoting the phase field used to track the fracture pattern. The cracked region is represented by $\phi = 1$ while the undamaged portion is given by $\phi = 0$. The fracture energy, $\Psi_c$, is defined in terms of the critical energy release rate, $G_c$, integrated over the fracture surface. The phase-field approximation introduces an $n$-th order crack density functional, $\Theta$, that is dependent only on a length scale parameter, $l_0$, the phase-
field, $\phi$, and derivatives of $\phi$ up to order $n$ such that the approximation stated in \autoref{eq:totalenergy} for the fracture energy holds true. A main feature of phase-field modeling is the assumption that the process zone has a finite width, which is controlled by $l_0$. A sharp crack topology is recovered in the limit as $l_0 \to 0$ \cite{Bourdin2000}. The external potential energy, $\mathcal{P}_{ext}$, is computed using the prescribed boundary force. $\bm{t_N}$ is the traction load applied over the Neumann boundary, $\partial\Omega_N$, and the distributed body force, $\bm{f}$, is applied over the whole domain. In this approach, the effects associated with crack formation such as stress release are incorporated into the constitutive model. 

In the phase field approach, the crack path is resolved by minimizing the energy functional, $\mathcal E$, defined in \autoref{eq:totalenergy}. $\Psi_e$ describes a smooth transition from the intact bulk material to the fully cracked state, characterized by $\psi_e\left(\bm{\epsilon}\right)$ and a monotonically decreasing stress-degradation function, $g(\phi)$, which reduces the stiffness of the bulk material. Taking into account that the compressive strain energy does not participate in the propagation of the crack, a tension-compression split of $\Psi_e(\epsilon)$ is considered as:
\begin{equation}\label{eq:strain_decomposed}
    \begin{split}
       \Psi_e\left(\bm{\epsilon}\right) &= g\left(\phi\right) \Psi_{e}^{+}\left(\bm{\epsilon}\right) + \Psi_{e}^{-}\left(\bm{\epsilon}\right),\\
        \text{where\;\;}g(\phi) &= (1-\phi)^2.
    \end{split}
\end{equation}
$\Psi_{e}^{+}$ and $\Psi_{e}^{-}$ are the tensile and the compressive components of the strain energies obtained by the spectral decomposition of the strain tensor.

Since the crack is modeled as a damage region, an exponentially decaying function is introduced to approximate the non-smooth crack topology. In one-dimension, to denote a crack located at $x=a$, a particular form of the second-order phase field model is given by \cite{Miehe2010}:
\begin{equation}\label{eq:crack_diffusion_2nd}
    \phi ({x}) = \exp\left(\frac{-| x-a|}{l_0}\right).
\end{equation}
Note that \autoref{eq:crack_diffusion_2nd} is the solution of the homogeneous ordinary differential equation \cite{Miehe2010}:
\begin{equation}\label{eq:1d_ODE_2nd}
    \phi''({x}) - \frac{1}{l_0^2} \phi ({x}) = 0 \text{ in } \Omega,
\end{equation}
subjected to the Dirichlet-type boundary conditions: 
\begin{equation}\label{eq:1D-dirichlet}
\begin{split}
    \phi\left(0\right) &= 1 \text{  , } \phi'\left(0\right) = 0 \text{  , }\\ \lim\limits_{ x\to\infty}\phi\left( x\right) &=  \lim\limits_{x \to-\infty}\phi\left( x\right)= 0,  \\
     \text{  and  } \lim\limits_{ x\to\infty}\phi'\left({x}\right) &=  \lim\limits_{x\to-\infty}\phi'\left(x\right)= 0.
    \end{split}
\end{equation}
\autoref{eq:1d_ODE_2nd} is the strong form associated with the variational problem:
\begin{equation}\label{eq:variational_problem}
    \phi^*  = \underset{\phi\in\mathbb{S}}{\text{argmin}}\, I(\phi),
\end{equation}
where 
\begin{equation}
\begin{split}
    &I(\phi) = \frac{1}{2}\int\limits_\Omega{\left(\phi^2 + l_0^2 |\nabla\phi|^2 \right)}d\Omega,\\
     \mathbb{S}:=&\{\phi(x)|\phi\left(0\right) = 1 \text{  and  } \lim\limits_{x\to\infty}\phi\left(x\right) =  \lim\limits_{x\to-\infty}\phi\left(x\right)= 0.\}
\end{split}
\end{equation}
As a consequence, $\Theta_2(\phi, l_0)$ for the second-order phase field model is defined as:
\begin{equation}\label{eq:fracture_density_2nd}
        \Theta_2 (\phi, l_0) = 
        \frac{1}{2l_0} \left(\phi^2 + l_0^2|\nabla\phi|^2 \right).
\end{equation}
For the second-order phase field model, $\Psi_c$ is approximated as
\begin{equation}\label{eq:fracture_energy_2nd}
    \Psi_c = \int_{\Gamma}{G_c}\;d\Gamma
    \approx\frac{G_c}{2l_0} \int\limits_\Omega{\left(\phi^2 +l_0^2 |\nabla\phi|^2 \right)}d\Omega.
\end{equation}
If the phase field is computed in case of loading, \autoref{eq:fracture_energy_2nd} is reformulated as:
\begin{equation}\label{eq:fracture_energy_loading}
    \Psi_c \approx \frac{G_c}{2l_0} \int\limits_\Omega{\left(\phi^2 +l_0^2 |\nabla\phi|^2 - g(\phi)\Psi_{0}^{+}\right)}d\Omega,
\end{equation}
along with proper irreversibility conditions applied on $\phi$ to guarantee that all cracks must always extend over time. Similar formulations can also be employed in 2D and 3D problems.

To enforce irreversibility, a strain history functional, $H(\bm{x},t)$ was introduced in \cite{Miehe2010a} and is defined as:
\begin{equation}\label{eq:history_field}
    H(\bm{x},t) = {\max_{s \in [0,t]}}\Psi^{+}_{0}(\bm{\epsilon}(\bm{x},s)),
\end{equation}
where $\bm{x}$ is the integration point. The strain history functional replaces $\Psi_{0}^{+}$ in \autoref{eq:fracture_energy_loading}. One advantage of using the history function is that it could be used to model pre-existing cracks in the domain. The initial strain history function, $H(\bm{x},0)$ could be defined in terms of $d(\bm{x},l)$, which is the closest distance from any point, $\bm{x}$ on the domain to the line, $l$ which represents the discrete crack. In particular, we set
\begin{equation}\label{eq:initial_history_field}
    H(\bm{x},0) = \left\{ {\begin{array}{l l}
  {\frac{BG_c}{2l_0}(1 - \frac{2d(\bm{x},l)}{l_0})}&{d(\bm{x},l) \leqslant \frac{l_0}{2}} \\ 
  0&{d(\bm{x},l) > \frac{l_0}{2}}
\end{array}} \right.,
\end{equation}
where $B$ is a scalar parameter that controls the magnitude of the scalar history field and is calculated as:
\begin{equation}\label{eq:scalarB}
    B = \frac{1}{1-\phi}  \; \; \; \text{for} \;\;\phi < 1.
\end{equation}
The phase-field is assumed to satisfy homogeneous Neumann-type boundary conditions on the entire boundary:
\begin{equation}\label{eq:Phase_boundary}
    \nabla\phi\cdot \bm{n} = 0 \text{ on } \partial\Omega,
\end{equation}
where $\bm{n}$ is the unit outward normal vector. In the energy method, the solution is obtained by minimization of the total variational energy of the system, $\mathcal{E}$. The problem statement can be written as:
\begin{equation}\label{eq:ProbStatement_energy}
\begin{split}
    \text{Minimize:}\;\;\;\; \mathcal{E} &= \Psi_e + \Psi_c,\\
\end{split}
\end{equation}
subject to proper boundary conditions. In \autoref{eq:ProbStatement_energy}, $\Psi_e$ is the stored elastic strain energy and $\Psi_c$ is the fracture energy. 
In this work, without loss of generality, we consider the Dirichlet condition on displacement, while the approach can also be extended to other types of conditions. Using the variational approach, the traction-free Neumann boundary conditions are automatically satisfied. 
In \autoref{eq:ProbStatement_energy}, $\Psi_e$ and $\Psi_c$ are defined as:
\begin{equation}\label{eq:energyterms}
    \begin{split}
       \Psi_e &=  \int_{\Omega}f_e(\bm{x}) d\Omega,\\
       \Psi_c &=  \int\limits_\Omega f_c(\bm{x}) d\Omega,\\
    \end{split}
\end{equation}
where
\begin{equation}\label{eq:indv_ent}
    \begin{split}
        f_e(\bm{x}) &=  g(\phi) \Psi_{0}^+\left(\bm{\epsilon}\right) + \Psi_{0}^-\left(\bm{\epsilon}\right),\\
       f_c(\bm{x}) &=  \frac{G_c}{2l_0} \left(\phi^2 +l_0^2 |\nabla\phi|^2 \right) - g(\phi)H(\bm{x},t).\\
    \end{split}
\end{equation}
We have used the monolithic solution scheme to solve the coupled fields, where $\Psi_e$ and $\Psi_c$ are simultaneously minimized (by directly minimizing $\mathcal E$) to obtain the displacement field and the phase-field. In all the brittle fracture problems, the crack is initialized using $H(\xb,0)$ defined in \autoref{eq:initial_history_field}. In the next section, we will provide an overview of the conventional DeepONet framework put forth in \cite{lu2021learning} before we discuss how the variational formulation is integrated in the DeepONet to develop a surrogate model for fracture analysis.

\section{DeepONet}
\label{sec:deeponet}

The idea of DeepONet is motivated by the universal approximation theorem for operators \cite{chen1995universal}, which states that a neural network with a single hidden layer can approximate accurately any linear/non-linear continuous function or operator. Before we focus on learning the solution operators of the parametric PDEs, it is important to understand the difference between a function regression and an operator regression. In the function regression approach, the solution operator is parameterized as a neural network between finite Euclidean spaces: $\mathcal{F}:\mathbb{R}^{d_1}\to\real^{d_1}$, where $d_1$ is the number of discretization points.
However, in operator regression, a function is mapped to another function through an operator. In other words, it is the mapping of infinite-dimensional space to another infinite dimensional space. Using operator regression, the operators would be trained to approximate the solution of the input functions by learning the non-linear operator from the data. 

\begin{figure}[htbp!]
    \centering
    \subfigure[]{
    \includegraphics[width = \textwidth]{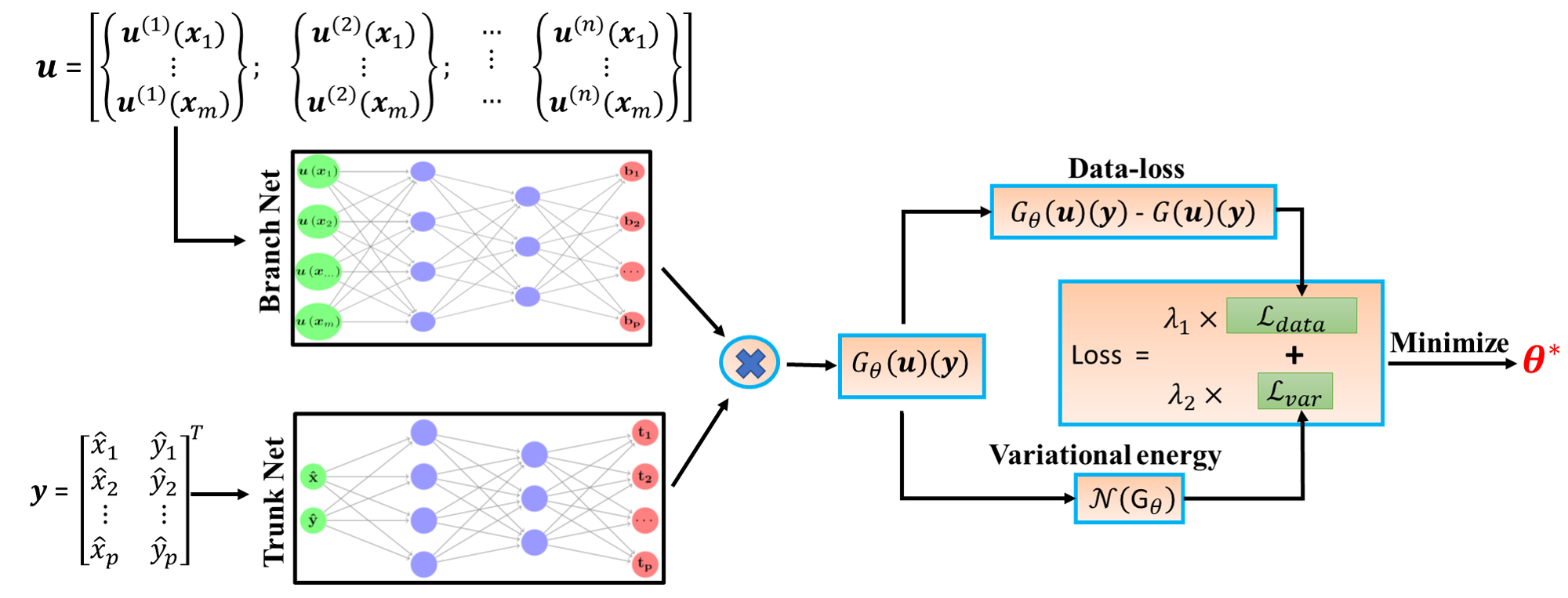}}
    \subfigure[]{
    \includegraphics[width = \textwidth]{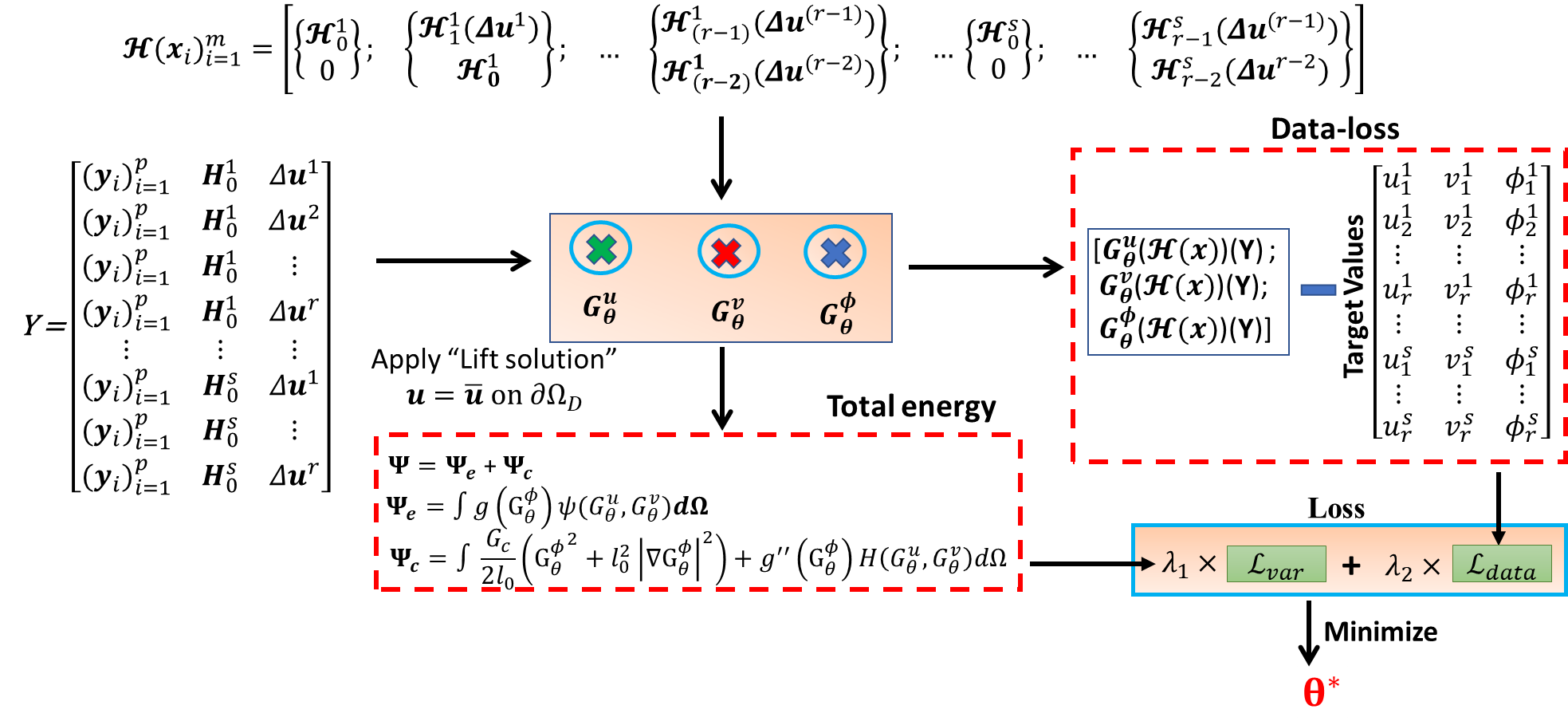}}
    \caption{(a) Schematic representation of DeepONet. The branch net takes the function $\{\bm u^{(i)}\}_{i=1}^n$ as input evaluated at $m$ fixed sensor points, denoted by $\{\xb_1, \xb_2, \ldots, \xb_m\}$ and returns features embedding $[b_1, b_2, \ldots, b_q]^T \in \real^q$ as output. The trunk net takes the continuous coordinates and parameters, $\bm y_i =\{\hat{x_i},\hat{y_i}\}_{i=1}^p$ $\in$ $\mathbf Y$ as inputs, and outputs a features embedding $[t_1, t_2, \ldots, t_q]^T \in \real^q$. The features embedding of the branch and trunk networks are merged via a dot product to output the solution operator, $\mathcal G_{\bm{\theta}}(\bm u)(\bm y)$. The parameter $\bm{\theta}$ denotes the collection of all trainable weights and bias in the branch and the trunk network. The optimized parameters, $\bm \theta^*$, are obtained by minimizing the hybrid loss function defined as the weighted sum of variational energy, $\mathcal L_{var}$, and the data-driven loss, $\mathcal L_{data}$, where the weights are denoted by $\lambda_1$ and $\lambda_2$. (b) The computational framework of the proposed surrogate model is shown schematically. The input to the branch net is the tensile strain energy of previous two displacement steps ($\bm{\mathcal H}_{r-1}, \bm{\mathcal H}_{r-2}$), while the input to the trunk net is the evaluation coordinates, $\bm y$, the initial configuration described using \autoref{eq:initial_history_field}, and the applied displacement. For predicting the crack located in the first displacement step, the input to the branch net is the initial strain energy, $\bm{\mathcal H}_0$, padded with zeros at all sensor locations to indicate no prior history.}
    \label{fig:unstacked_DeepONet}
\end{figure}

The DeepONet architecture consists of two neural networks: one encodes the input function, $\boldsymbol u$ at fixed sensor points (branch net), while another represents the output for the location, $\boldsymbol y$ of evaluation of the output function (trunk net). The success of deep learning has been largely attributed to the depth of the networks, i.e., the number of successive affine transformations followed by non-linearity, which is shown to be extracting hierarchical features from the data. Hence, in this work, we have used two deep, fully connected feed-forward neural networks (branch net and trunk net) to approximate the solution operator. The goal of the DeepONet algorithm is to learn the operator, $\mathcal G$, which takes as an input the function $\boldsymbol u$ in the branch net, and then $\mathcal G(\boldsymbol u)$ is the corresponding output function. The output of the branch net is evaluated at $\boldsymbol y$ continuous coordinates (input to the trunk net). Although the architecture proposed here can be applied to more general problems, in the following we illustrate the formulation on a 2D problem for simplicity. The output of the DeepONet is a scalar and is expressed as $\mathcal G_{\bm\theta}(\boldsymbol u)(\boldsymbol y)$, where $\bm{\theta} = \left(\mathbf W, \bb \right)$ includes the trainable parameters (weights, $\mathbf W$, and biases, $\bb$) of the DeepONet. The input functions to the branch net may include, the shape of the physical domain, the initial or boundary conditions, constant or variable coefficients, source terms, etc. Even though the branch net takes a function as input, we have to represent the input functions discretely, so that network approximations can be applied. To that end, all the input functions, $\boldsymbol u$, are evaluated at finite locations, $\mathcal X=\{\bm x_1, \bm x_2, \ldots, \bm x_m\}$, referred to as sensors. The location of the $m$ sensors must be the same for all the functions, $\boldsymbol u$. We do not enforce any constraints on the output locations, $\bm y$. A schematic representation of  DeepONet is shown in \autoref{fig:unstacked_DeepONet}(a), where the branch net takes as input $n$ functions represented as $\boldsymbol u^{(1)}, \boldsymbol u^{(2)}, \boldsymbol u^{(3)}, \ldots, \boldsymbol u^{(n)}$, and the solution operator is evaluated at $\bm y = \{\yb_1,\yb_2,\cdots,\yb_p\}=\{(\hat x_1,\hat y_1),(\hat x_2, \hat y_2), \ldots, (\hat x_p,\hat y_p)\}$, which are the inputs to the trunk net. Here $\hat{x}_i$ $\hat{y}_i$ denote the $x$ and $y$ coordinates of point $\yb_i$, respectively.

Let us consider that the branch neural network consists of $l_{br}$ hidden layers, where the $l_{br}$-th layer is the output layer consisting of $q$ neurons. Considering an input function, $\bm u^{(i)}$ in the branch net, the network returns a feature embedding $[b_1, b_2, \ldots, b_q]^T$ as output. The output, $\bm{Z}_{br}^{l_{br}}$ of the feed-forward branch neural network is expressed as:
\begin{equation}\label{eq:output_branch}
    \begin{split}
    \bm{Z}_{br}^{l_{br}} &= \left[b_1, b_2, \ldots, b_q\right]^T,\\
    &=\sigma_{br}\left(\mathbf W^{l_{br}}\bm{z}^{l_{br}-1} + \bb^{l_{br}}\right),
    \end{split}
\end{equation}
where $\sigma_{br}\left( \cdot \right)$ denotes the non-linear activation function for the branch net and $\bm{z}^{l_{br}-1} = f(\bm u^{(i)}(\bm x_1), \bm u^{(i)}(\bm x_2), \ldots, \bm u^{(i)}(\bm x_m))$, where $f\left(\cdot\right)$ denotes a function. Similarly, consider a trunk network with $l_{tr}$ hidden layers, where the $l_{tr}$-th layer is the output layer consisting of $q$ neurons. The trunk net takes the continuous coordinates in $\bm y$ as inputs, and outputs a features embedding $[t_1, t_2, \ldots, t_q]^T$. The output of the trunk net can be represented as:
\begin{equation}\label{eq:output_trunk}
    \begin{split}
    \bm{Z}_{tr}^{l_{tr}} &= \left[t_1, t_2, \ldots, t_q\right]^T,\\
    &=\sigma_{tr}\left(\mathbf W^{l_{tr}}\bm{z}^{l_{tr}-1} + \bb^{l_{tr}}\right),
    \end{split}
\end{equation}
where $\sigma_{tr}\left( \cdot \right)$ denotes the non-linear activation function for the trunk net  and $\bm{z}^{l_{tr}-1} = f(\bm y_1, \bm y_2, \ldots, \bm y_p)$, where $\bm y_j = \{(\hat x_j, \hat y_j)\}_{j=1}^p$. For a single input function, $\bm u^{(i)}$, the DeepONet prediction, $\mathcal G_{\bm \theta}(\bm u)$, evaluated at any coordinate, $\bm y$ can be expressed as:
\begin{equation}\label{eq:output_deeponets}
    \begin{split}
    \mathcal G_{\bm{\theta}}(\bm u^{(i)})(\bm y) &= \sum_{k = 1}^{q}\left(\sigma_{br}(\mathbf W^{l_{br}}_k\bm{z}^{l_{br}-1}_k + \bb^{l_{br}}_k)\cdot \sigma_{tr}(\mathbf W^{l_{tr}}_k\bm{z}^{l_{tr}-1}_k + \bb^{l_{tr}}_k)\right),\\
    &= \sum_{k = 1}^{q}b_k(\bm u^{(i)}(x_1), \bm u^{(i)}(x_2), \ldots, \bm u^{(i)}(x_m))\cdot t_k(\bm y).
    \end{split}
\end{equation}

In general, a DeepONet training dataset is a triplet of the form, $\left[\{\bm u^{(i)}\}_{i=1}^n, \{\bm y_j\}_{j=1}^p, \mathcal G(\bm u)(\bm y)\right]$:
\begin{equation}\label{eq:triplet}
    \begin{bmatrix}
    \begin{bmatrix}
    \bm u^{(1)}(\bm x_1), \bm u^{(1)}(\bm x_2), \ldots, \bm u^{(1)}(\bm x_m)\\
    \bm u^{(1)}(\bm x_1), \bm u^{(1)}(\bm x_2), \ldots, \bm u^{(1)}(\bm x_m)\\
    \vdots\\
    \bm u^{(1)}(\bm x_1), \bm u^{(1)}(\bm x_2), \ldots, \bm u^{(1)}(\bm x_m)\\
    \bm u^{(i)}(\bm x_1), \bm u^{(i)}(\bm x_2), \ldots, \bm u^{(i)}(\bm x_m)\\
    \vdots\\
    \bm u^{(n)}(\bm x_1), \bm u^{(n)}(\bm x_2), \ldots, \bm u^{(n)}(\bm x_m)\\
    \end{bmatrix} ,& \begin{bmatrix}
    \bm y_1^{(1)}\\
    \bm y_2^{(1)}\\
    \vdots\\
    \bm y_p^{(1)}\\
    \bm y_1^{(i)}\\
    \vdots\\
    \bm y_p^{(n)}    
    \end{bmatrix} ,& \begin{bmatrix}
    \mathcal G(\bm u^{(1)})(\bm y_1^{(1)})\\
    \mathcal G(\bm u^{(1)})(\bm y_2^{(1)})\\
    \vdots\\
    \mathcal G(\bm u^{(1)})(\bm y_p^{(1)})\\
    \mathcal G(\bm u^{(i)})(\bm y_1^{(i)})\\
    \vdots\\
    \mathcal G(\bm u^{(n)})(\bm y_p^{(n)})
    \end{bmatrix}\\
    \end{bmatrix}.
\end{equation}
In the triplet shown in \autoref{eq:triplet}, each input function, $\bm u^{(i)}$ is repeated $p$ times, where $p$ is the number of points at which the solution operator, $\mathcal G_{\bm{\theta}}(\bm u^{(i)})$, is evaluated to construct the loss function. The training dataset of a DeepONet consists of three parts; input to the branch net, $\bm u$, input to the trunk net, $\bm y$, and the target values of the solution, $\mathcal G(\bm u)(\bm y)$. Taking a two-dimensional problem with scalar-valued input $\ub(\xb)$ and output $\mathcal G(\bm u)(\bm y)$ for illustration, the tensor dimensions for each of the components of the training set are: $\text{dim}(\bm u) := (n\times p,m)$, $\text{dim}(\bm y) := (n\times p,2)$, and $\text{dim}(\mathcal G(\bm u)(\bm y)) := (n\times p,1)$.  DeepONet requires large annotated data-sets consisting of paired input-output observations, while they provide a simple and intuitive model architecture that is fast to train, allowing for a continuous representation of the target output functions that is independent of resolution. In \autoref{eq:triplet}, $\mathcal G(\bm u)(\bm y)$ represents the ground truth, which could be obtained either from experimental data or from high-fidelity simulations. Conventionally, the trainable parameters of the DeepONet represented by $\bm{\theta}$ in \autoref{eq:output_deeponets} is obtained by minimizing a loss function. Common loss functions used in literature includes the $l_2$-loss function and the $l_1$-loss function \cite{Rojas1996neural}. 
\begin{equation}\label{eq:L1_loss}
\begin{split}
    l_1 &= \sum_{i =1}^n \sum_{j =1}^p \left| \mathcal G(\bm u^{(i)})(\bm y_j^{(i)}) - \mathcal G_{\bm{\theta}}(\bm u^{(i)})(\bm y_j^{(i)})\right|,\\
    l_2 &= \sum_{i =1}^n \sum_{j =1}^p\left(\mathcal G(\bm u^{(i)})(\bm y_j^{(i)}) - \mathcal G_{\bm{\theta}}(\bm u^{(i)})(\bm y_j^{(i)})\right)^2,\\
\end{split}
\end{equation}
where $\mathcal G_{\bm{\theta}}(\bm u^{(i)})(\bm y_j^{(i)})$ is the predicted value obtained from the DeepOnet, while  $\mathcal G(\bm u^{(i)})(\bm y_j^{(i)})$ is the target value.

DeepONet has shown remarkable success in diverse fields of applications like electro-convection multiphysics, bubble dynamics, inelastic impact problems,  etc., where the network is trained using large datasets. However, in fracture mechanics,  collecting a large amount of data from experiments is improbable as performing experiments for various different crack lengths, locations, and material parameters would render the process very expensive. Moreover, it is infeasible to perform a rigorous and computationally intensive crack-growth simulation within the possible short time span following a discrete-source damage event. Hence, in a small data regime, we will encode the physical laws that govern the growth of fracture to train the DeepONet for predicting damage paths. Since the growth of fracture is energy-driven, we encode the variational form of the governing PDE into the DeepONet, terming it the variational energy-based DeepONet (V-DeepONet). Along with the physics of the problem, we use relatively small input-output datasets to improve the prediction accuracy of the network, thereby proposing a hybrid loss function. In the next section, we present the V-DeepONet algorithm to compute the optimized parameters, $\bm \theta^*$, associated with the two deep neural network architecture. 

\section{Variational energy based DeepONet}
\label{sec:vedeeponet}

In this section, we develop the V-DeepONet, which is inspired by the variational form of the governing PDE. Without the loss of generality, we consider the physics of a problem, defined by a generic time-independent differential equation of the form:
\begin{subequations}\label{eq:ps}
\begin{equation}\label{eq:generic_de}
    \mathcal K \left(\bm w,\nabla \wb, \ldots,\nabla^{\alpha} \wb,
    \bm x, \fb\left(\bm x\right) \right) = 0,\;\;\; \xb \in \Omega, 
\end{equation}
\vspace*{-\baselineskip}
\begin{equation}\label{eq:gen_bc}
    \bm w\left(\bm x_D\right) = \bm w_D,\;\;\; \bm x_D \in \partial \Omega, 
\end{equation}
\end{subequations}
defined over the physical domain, $\Omega$ with boundaries, $\partial \Omega$. 
$\nabla\wb$ denotes the first order derivatives of $\wb$ with respect to $\xb$, and $\nabla^\alpha \wb$ represents all derivatives of $\wb$ with the form $\frac{\partial^{\alpha}\wb_i}{\partial \xb_1^{\alpha_1}\cdots\xb_d^{\alpha_d}}$ with $\sum_{i=1}^d{\alpha_i}=\alpha$, where $\wb_i$ is the i-th component of $\wb$ and similarly for $\xb_i$. $\alpha$ denotes the highest order of derivative required to describe the underlying PDE. The forcing function, $\fb(\bm x)$ is a known source term and the operator, $\mathcal K$ is usually a differential or integro-differential operator. \autoref{eq:gen_bc} represents the Dirichlet boundary condition, where $\bm x_D$ represents a Dirichlet boundary point. Since the method is based on the energy principle, the homogeneous Neumann boundary conditions are automatically satisfied. Assuming that $\fb(\bm x)$ is known, we aim to learn the solution operator such that:
\begin{equation}
    \mathcal G_{\bm{\theta}}: f(\bm x) \rightarrow \bm w(\bm x).
\end{equation}
Based on the formulation of DeepONet defined in \autoref{sec:deeponet}, the input function of the branch net is the source term, $f(\bm x)$, which is evaluated at $\{\bm x_i\}_{i=1}^{m}$ sensor points, where $\bm x_i \in \mathcal X$. The DeepONet is approximating the solution operator, $\mathcal G_{\bm{\theta}}\left(\fb\left(\bm x\right)\right)$, which is evaluated at a set of points, $\bm y_{j=1}^p$, that are randomly sampled in the domain of $G_{\bm{\theta}}\left(\fb\left(\bm x\right)\right)$, and are used to approximately enforce a set of given physical constraints, typically described by the PDE in \autoref{eq:generic_de}. Let the variational energy formulation of \autoref{eq:generic_de} be expressed as:
\begin{equation}\label{eq:tot_energy}
    \mathcal V_e = \int_{\Omega}\mathcal F \left(\bm w,\nabla\bm w,\ldots, \nabla^\alpha\bm w, \bm x, \fb\left(\bm x\right) \right) d\Omega,
\end{equation}
where $\mathcal F$ is a differentiable functional. With this, the solution to \autoref{eq:generic_de} can be obtained by solving the following optimization problem:
\begin{equation}\label{eq:opt_ve}
\begin{split}
    & \bm w^* = \argmin_{w} \mathcal V_e \left(\mathcal G(\fb(\bm x)) \right)\\
    \text{subject to:}\;\; &\mathcal G_{\bm \theta}(\fb(\bm x))\left(\bm y = \bm x_D\right)= \bm w_D.
\end{split}
\end{equation}
In V-DeepONet, we utilise the same approach as discussed in \autoref{eq:output_deeponets} to obtain the solution operator, $\mathcal G_{\bm \theta}(\fb(\bm x))$, with parameters, $\bm{\theta} =\left[\mathbf W, \bb \right]$. Next, the DeepONet outputs are modified in such a way so that the solution operator when evaluated at the Dirichlet boundary points, the boundary conditions are exactly satisfied. Therefore, the modified output of the DeepONet is defined as:
\begin{equation}\label{eq:auto_bc}
    \mathcal G_{\bm{\theta}}(f(\bm x))\left(\bm y\right) = \tilde{\bm w}_D + B(\bm y)\cdot\hat{\mathcal G}_{\bm{\theta}}(\fb(\bm x))\left(\bm y\right),
\end{equation}
where $\hat{\mathcal G}_{\bm{\theta}}$ is the solution obtained from the DeepONet, $\tilde{\bm w}_D$ is a function chosen such that $\tilde{\bm w}_D = \bm w_D$ and $B(\bm y)=0$ on the Dirichlet boundary \cite{weinan2018deep}. Hence, the boundary conditions are satisfied and we have no boundary-loss term in the loss function.
In the next step, we compute the derivatives of the solution operator, $\mathcal G_{\bm{\theta}}(\fb(\bm x))\left(\bm y\right)$, with respect to the spatial co-ordinates, $\left(\hat{x},\hat{ y}\right)$, defined over the domain using the automatic differentiation technique. These derivatives are components of the variational energy formulation stated in \autoref{eq:tot_energy}. The computed derivatives are substituted in \autoref{eq:tot_energy} along with the solution operator approximating $\bm w (\bm x)$ to obtain the total energy. The network parameters can be trained by minimizing the hybrid loss function, $\mathcal L\left(\bm \theta\right)$, which is defined as:
\begin{equation}\label{eq:DeepOnet_loss}
\begin{split}
    \mathcal L\left(\bm \theta\right) &= \lambda_1 \times \mathcal L_{data}\left(\bm{\theta}\right) + \lambda_2 \times \mathcal L_{var}\left(\bm{\theta}\right),\\
    \mathcal L_{data}\left(\bm \theta\right) &= \frac{\sum\limits_{i =1}^n \sum\limits_{j =1}^m \sum\limits_{k =1}^p\left(\mathcal G_{\bm \theta}(f^{(i)}(\bm x_j))(\bm y_k)- \mathcal G(f^{(i)}(\bm x_j))(\bm y_k)\right)^2}{n\times p \times m},\\  
    \mathcal L_{var}\left(\bm \theta\right) &= \frac{\sum\limits_{i =1}^n \sum\limits_{j =1}^p \sum\limits_{k =1}^m \mathcal F \left(\bm w,\nabla\bm w,\ldots, \nabla^\alpha\bm w, \bm x, f\left(\bm x\right) \right)}{n\times p \times m},
\end{split}
\end{equation}
where $\bm w\left(\bm x\right)\approx \mathcal G_{\bm{\theta}}(\fb^{(i)}(\bm x_j))$ and $\lambda_1$ and $\lambda_2$ are the weights pertaining to each of the component of the loss function. The developed approach seamlessly integrates the data measurements and variational form of the PDE by penalizing the total energy in the hybrid loss function of the V-DeepONet.

In this work, we aim to train the V-DeepONet such that it could be used as a surrogate model to predict the damage path for any given initial defect and for any applied displacement (considering displacement-controlled fracture). To design a surrogate model in the framework of V-DeepONet, we need to choose the input functions, sensor points, parametrized variables judiciously and strike a balance between efficiency and accuracy of the network. 

\section{Surrogate modeling for fracture analysis}
\label{sec:surrogate_fracture}

In this section, we provide a detailed description of the proposed surrogate model to infer the probable failure paths. The solution of the coupled problem consists of a $d$-dimensional vector-valued displacement field and a scalar-valued phase field output. In all the examples presented in this paper, we have considered two-dimensional problems, hence $d=2$. Therefore, for each material point $\yb$, the output of the V-DeepONet has three components. Noticing that the output in the vanilla DeepONet \autoref{eq:triplet} is a scalar, the neural network output architecture needs to be modified. To resolve this issue, we modify the forward pass in \autoref{eq:output_deeponets} such that the output of the V-DeepONet can be a vector. The tensor dimensions of $\mathcal G(\bm u)(\bm y)$ in \autoref{eq:triplet} will be $(n\times p, 3)$. The employed V-DeepONet is defined as $\mathcal G_{\bm{\theta}} = \left[\mathcal G_{\bm{\theta}}^{u}, \mathcal G_{\bm{\theta}}^{v}, \mathcal G_{\bm{\theta}}^{\phi}\right]$ to represent the solution map from initial conditions to the associated solutions, where $\mathcal G_{\bm{\theta}}^{u}$, $\mathcal G_{\bm{\theta}}^{v}$ are the solution for the displacement components along $x$-direction and $y$-direction, respectively and $\mathcal G_{\bm{\theta}}^{\phi}$ is the solution for the phase field, $\phi$. 

The goal of the V-DeepONet based surrogate model is to find the possible pathways of failure based on the initial crack configuration. Failure of brittle materials typically starts with the development of a region of high stress around the crack tip, and our method will be employed based on the locations of the regions of high-stress. In this method, we assume the presence of initial defects when the system is not loaded (initial stage). The initial history function defined in \autoref{eq:initial_history_field} provides a simple, mesh independent mechanism for adding initial defects to a model. Using this approach, the initial crack is identified by a region of very high strain energy, whereas the strain energy is close to zero in other parts of the domain. The proposed method focuses on displacement-controlled failure. As the system is loaded (prescribed displacement is increased incrementally), pre-existing cracks grow leading to the formation of new edges and finally leading to the failure of the system. 

To understand how the surrogate model works, we divide the aim of the surrogate model into two sub-parts in order:
\begin{itemize}
    \item To predict the crack location and the deformed configuration of the domain for any applied displacement, given a fixed initial condition.
    \item To predict the final crack path for any location of the initial defect, given a fixed applied displacement.
\end{itemize}
In this section, we will develop independent surrogate models to meet each of the two goals mentioned above. The final surrogate model will then be engineered as an integration of the above two surrogate models. In \autoref{subsec:surrogate1}, we focus on constructing a surrogate model to predict the crack location for any applied displacement, when the domain geometry and the initial condition are fixed. Then in \autoref{subsec:surrogate2}, we will discuss the surrogate model for predicting the crack path for any pre-crack location, with a fixed displacement loading applied on the boundary.

\subsection{Surrogate 1: To predict the crack location for any applied displacement}
\label{subsec:surrogate1}

In this section, we present our first contribution for solving displacement-controlled quasi-static fracture problem. At each step, we update the displacement boundary condition with an increment as $\wb_D\to\wb_D+\Delta \wb$, then solve the nonlinear optimization problem defined in \autoref{eq:opt_ve} with this new boundary condition applied. In conventional numerical solvers, small displacement increments $\Delta \wb$ are essential to capture the brutal nature of crack growth. V-PINNs, which were proposed in \cite{goswami2020transfer}, successfully implemented the crack growth problem using larger displacement increments and enabled a significant reduction in the computational cost. However, there is a major bottleneck of V-PINNs: the fracture problem is sequentially solved as a short loading-step problem with independent neural network training for each loading step. Even though the concept of transfer learning was implemented in the framework of V-PINNs, the method still remained computationally expensive.

In this work, we train a single V-DeepONet to learn the solution operator of the same PDE for small displacement steps subject to same initial condition. The steps involved in the proposed approach are as follows:
\begin{itemize}
\item First, we decide the location of the $m$ sensor points to distinguish two input functions (defined in terms of tensile strain energy) of the branch net.
\item Next, for training the V-DeepONet, 
we consider $n$ steps with corresponding displacement increments, denoted by $\Delta \wb_1, \Delta \wb_2, \ldots, \Delta \wb_n$. Therefore, the input to the branch net will be $n$ tensile strain energy evaluated at $m$ sensor locations. The input function corresponding to the displacement step $\Delta \wb_{i}$ will be the tensile strain energy due to an applied displacement $\Delta \wb_{i-1}$, where $i \in \{2,\cdots, n\}$. Here, the tensile strain energy is obtained from the results of high-fidelity IGA simulations. To predict the solution fields for an applied displacement of $\Delta \wb_1$, the tensile strain energy computed using \autoref{eq:initial_history_field} at the sensor locations is applied as the input to the branch net. The solution operators are defined such that:
\begin{equation}
    \left[\mathcal G_{\bm{\theta}}^{u}, \mathcal G_{\bm{\theta}}^{u}, \mathcal G_{\bm{\theta}}^{\phi}\right]: \bm{\mathcal H}(\bm y) \rightarrow \left[ u(\bm y), v(\bm y), \phi(\bm y)\right],
\end{equation}
where $\bm{\mathcal H}(\bm y)$ is the tensile strain energy, and $u$, $v$ denote the $x-$ and $y-$component of the displacement field, respectively.
\item In the third step, we prepare the inputs to the trunk net for extracting latent representations of the input coordinates. For each displacement step $\Delta \ub_{i}$, $\{\bm y^{(i)}_j\}_{j=1}^p$ are a set of points sampled in the domain, $\Omega$. The output function is evaluated at $\bm y_j^{(i)}$ for an applied displacement, $\Delta \ub_i$, where $i \in \{1,\cdots,n\}$ .  
\item Next, we modify the elastic field outputs of the V-DeepONet, $\mathcal G_{\bm{\theta}}^{u}$ and $\mathcal G_{\bm{\theta}}^{v}$ as discussed in \autoref{eq:auto_bc} so that the Dirichlet boundary conditions are satisfied.
\item Finally, we construct the hybrid loss function as defined in \autoref{eq:DeepOnet_loss} and minimize the loss to obtain the optimized parameters, $\bm{\theta}^*$.
\end{itemize}
Once the V-DeepONet is trained, it can be used to predict the displacement field and the phase field for any applied displacement on a fixed initial condition. In order to predict the solution of the V-DeepONet at the $k$-th displacement step, the branch net takes as input the tensile energy at the sensor points obtained from the $(k-1)$-th displacement step. To obtain the tensile energy, we compute the displacement gradients, and the eigenvalues of the strain, $\lambda_1^E$ and $\lambda_2^E$ using the outputs of the V-DeepONet obtained for the $k-1$-th step at the sensor locations. The computed eigenvalues are then used to obtain $\Psi^+$ and $\Psi^-$.
\begin{subequations}\label{eq:decomposed_strain}
\begin{equation}
    \Psi^+ = \frac{\nu}{8}\left( \lambda_s + \left| \lambda_s \right| \right)^2 + \frac{\mu}{4}\sum_{i=1}^d\left( \lambda_i^E + \left| \lambda_i^E \right| \right)^2,
\end{equation}
\begin{equation}
    \Psi^- = \frac{\nu}{8}\left( \lambda_s - \left| \lambda_s \right| \right)^2 + \frac{\mu}{4}\sum_{i=1}^d\left( \lambda_i^E - \left| \lambda_i^E \right| \right)^2,
\end{equation}
\end{subequations}
where $\lambda_s = \sum_{i=1}^d \lambda_i^E$, and $\nu$ and $\mu$ are the Lam\'e constants. The $\Psi^+$ values at $m$ sensor locations is the input to the branch net for the $k$-th displacement step. 
To provide an implementation guidance for interested readers, in the following we provide tensor dimensions of the training set, in this surrogate, for each of the components in the triplet discussed in \autoref{eq:triplet}. In the triplet, $\text{dim}(\bm u) := (n\times p,m)$ corresponding to $n$ displacement steps, $p$ sampled points for the evaluation of the output function and $m$ fixed sensor points, $\text{dim}(\bm y) := (n\times p,3)$, where the first two columns correspond to the spatial location of the evaluation points and the third column corresponds to the applied displacement. In particular, the displacement increment in the principle direction is employed in as the third column of $\bm y$, as will be explained further in \autoref{sec:numericals}. Lastly, $\text{dim}(\mathcal G(\bm u)(\bm y)) := (n\times p,3)$, where the three columns are designated for the solution of $u$, $v$, and $\phi$, respectively.  

Having built the surrogate model for predicting the crack location for any applied displacement on a fixed initial condition, we now extend the context to training the surrogate model adequately over a wide range of potential crack starting locations. 

\subsection{Surrogate 2: To predict the final crack path for any initial crack location}
\label{subsec:surrogate2}

In this section, we train the V-DeepONet adequately over multiple initial conditions (initial location of the defects), to obtain the final crack path at fixed applied displacement, $\Delta \wb$. This surrogate model defines a mapping between the initial configuration defined using $\bm{\mathcal H}_0(\bm y)$ to the solution fields for a fixed applied displacement, $\Delta \wb$. The steps involved in the proposed approach are as follows: 
\begin{itemize}
    \item First, we identify $m$ sensor locations that could be used for adequate and distinct identification of different initial configuration. As mentioned previously,
    the location of the sensors must be the same for all input functions in the branch net.
    \item Next, we consider $n$ initial locations of the crack. For all the initial configurations, we compute the initial strain energy, $\{\bm{\mathcal H}_0^{(1)}, \bm{\mathcal H}_0^{(2)}, \ldots, \bm{\mathcal H}_0^{(n)}\}$ at the fixed sensor locations using \autoref{eq:initial_history_field}. The computed $\bm{\mathcal H}_0^{(i)}$ is the input function to the branch net.
    The tensor dimension of the input to the branch net will be $(n\times p,m)$. The solution operators are defined such that:
    \begin{equation}
        \left[\mathcal G_{\bm{\theta}}^{u}, \mathcal G_{\bm{\theta}}^{u}, \mathcal G_{\bm{\theta}}^{\phi}\right]: \bm{\mathcal H}_0(\bm y) \rightarrow \left[u(\bm y), v(\bm y), \phi(\bm y)\right].
    \end{equation}
    \item In the third step, we prepare the inputs to the trunk net for encoding the locations of the output function. For each of the initial condition, $\{\bm y^{(i)}_j\}_{j=1}^p$ is a set of points sampled in the domain, $\Omega$. The initial strain energy function is computed at the sampled points and is represented by $\bm H_j^{(i)}$. The tensor dimensions of the input to the trunk net will be $(n\times p,3)$, where the first two columns correspond to the spatial location of $\bm y_j$, and the third column provides the strain energy, $\bm H_j$, at the respective locations. The V-DeepONet is evaluated for $n$ initial conditions at all the sampled points in the trunk net.
    \item In the next step, the V-DeepONet outputs are modified to eliminate the boundary loss term from the loss function. Finally, the hybrid loss function is obtained and minimized to get the optimized parameters of the V-DeepONet.
\end{itemize}
Once this surrogate model is trained, it can be used to predict the displacement field and the phase field for any initial crack location at a fixed applied displacement. While testing the surrogate model, the analytically obtained history field (using \autoref{eq:initial_history_field}) at fixed sensor locations is employed as inputs to the branch net, while the evaluation coordinates and the initial strain energy (corresponding to the evaluation points) provide inputs of the trunk net.

Having discussed the surrogate models for crack locations at various displacement steps (with fixed initial condition) and final crack path for any initial crack location (with fixed applied displacement), we now shift the focus to develop a single surrogate model that can be trained to obtain the crack location at any applied displacement for any initial location of the crack.

\subsection{A unified model: To predict the crack location for any initial condition and any applied displacement}
\label{subsec:surrogate3}

Our final aim is to build a surrogate model such that the crack path can be traced from the initial configuration of the defect. This surrogate model would primarily be an integration of the previously discussed surrogate models. However, building such a surrogate model has two major challenges:
\begin{itemize}
    \item In \autoref{subsec:surrogate1}, to obtain the crack location at any given displacement step, we used the tensile strain energy only from the previous step. This approach is feasible only when a single initial crack configuration is considered. For multiple crack locations, there can be overlap of crack paths, and hence retaining the information of the initial crack location and/or multiple previous steps is essential. 
    \item To obtain the crack path for any initial location at any applied displacement, training the model using just the strain-energy of the previous displacement step (Surrogate 1) or the energy field in the vicinity of the original defect (Surrogate 2) is not sufficient, as the sequence or the original configuration might be lost, thereby leading to erroneous predictions. 
\end{itemize}
To resolve these challenges, we need to construct a V-DeepONet based surrogate model, which is capable of learning order dependencies in a sequence of prediction problems. Both these challenges will be addressed in the surrogate model, as will be discussed in this section. The following steps are considered for constructing the unified surrogate model:
\begin{itemize}
    \item Choice of the $m$ sensor locations: The location of the sensors should be carefully chosen such that they are representative of various initial configurations as well as crack locations at various applied displacement steps. It is worth emphasising that the input function space should be large enough to cover as many potential states of the underlying PDE system as possible. Otherwise, the trained model may not generalize very well for out-of-distribution initial conditions, possibly leading to large errors or even erroneous predictions.
    \item Next, we consider $s$ initial locations of the crack. For each initial condition, we compute the initial strain energy, $\{\bm{\mathcal H}_0^{(1)}, \bm{\mathcal H}_0^{(2)}, \ldots, \bm{\mathcal H}_0^{(s)}\}$, at the sensor locations using \autoref{eq:initial_history_field}. Now, for each initial crack configuration, we consider $r$ displacement steps, $\{\Delta \wb_1, \Delta \wb_2, \ldots, \Delta \wb_r\}$. Corresponding to every initial condition, we obtain the tensile strain energy, $\bm{\mathcal H}_{i}^{(j)}$, from the results of the IGA simulations, corresponding to an applied displacement $\Delta \wb_{i}$, where $i \in \{1,\cdots,r-1\}$ and $j\in \{1,\cdots,s\}$.  
    \item In the third step, we prepare the inputs for the branch net. To capture each sequential crack growth for $r$ applied displacements, we create a window of two steps, such that:
    \begin{equation}\label{eq:hist_window}
        \bm{\mathcal H} = \begin{bmatrix}
    \begin{bmatrix}
    \bm{\mathcal H}_0^{(1)}\\
    0
    \end{bmatrix} ;& \begin{bmatrix}
    \bm{\mathcal H}_1^{(1)}\\
    \bm{\mathcal H}_0^{(1)}
    \end{bmatrix} ;& \ldots, & \begin{bmatrix}
    \bm{\mathcal H}_{r-1}^{(1)}\\
    \bm{\mathcal H}_{r-2}^{(1)}
    \end{bmatrix} ;& \begin{bmatrix}
    \bm{\mathcal H}_{0}^{(2)}\\
    0
    \end{bmatrix} ;
    \ldots&;\begin{bmatrix}
    \bm{\mathcal H}_{r-1}^{(s)}\\
    \bm{\mathcal H}_{r-2}^{(s)}
    \end{bmatrix}\\
    \end{bmatrix}.
    \end{equation}
    The solution operators are defined as:
    \begin{equation}
        \left[\mathcal G_{\bm{\theta}}^{u}, \mathcal G_{\bm{\theta}}^{v}, \mathcal G_{\bm{\theta}}^{\phi}\right]: \bm{\mathcal H}(\bm y) \rightarrow \left[u(\bm y), v(\bm y), \phi(\bm y)\right],
    \end{equation}
    where $\bm{\mathcal H}(\bm y)$ is defined in \autoref{eq:hist_window}. The tensor dimensions of the input to the branch net will be $(r \times s\times p,2m)$, where the $r \times s = n$, the total number of input functions.
    \item In the next step, we prepare the input to the trunk net. To address the issue of retaining the original crack configuration for the associated branch net entries, the initial strain energy is included in the trunk net. For each initial crack location $\{\bm{\mathcal H}_0^{(i)}\}_{i=1}^s$, $\{\bm y^{(i)}_j\}_{j=1}^p$ is a set of points sampled in the domain, $\Omega$. The initial history function for $s$ initial conditions is computed at the sampled points and denoted as $\bm H_j^{(i)}$. The tensor dimensions of the input to the trunk net will be $(n\times p,4)$, where the first two columns correspond to $\bm y_j$, the third column is allocated for the history value, $\bm H_j$, and the fourth column is for the applied displacement. The V-DeepONet is evaluated at the locations and conditions defined in the trunk net.
    \item In the final step, the V-DeepONet outputs are modified to match the Dirichlet boundary condition. Then, the hybrid loss function is constructed and minimized to obtain the optimized parameters of the networks.
\end{itemize}
Once this model is trained, it can be used to make predictions for any initial configuration at various applied displacements sequentially. To predict crack paths using this model, we follow similar steps as discussed in \autoref{subsec:surrogate1}, except that in this surrogate we provide the strain energy of the last two displacements steps at the sensor location as the input to the branch net. \autoref{fig:unstacked_DeepONet}(b) provides a pictorial description of the framework and implementation details of the surrogate model to obtain the failure paths. For more clarity, an algorithm summarizing the overall framework is presented in Algorithm \ref{alg:V-deepONet_surr} of \autoref{app:appendixB}.

\section{Data Generation}
\label{sec:data_generation}

In \autoref{sec:surrogate_fracture}, we presented details about the proposed surrogate model developed within the framework of V-DeepONet. The model is trained using a hybrid loss function, which is the weighted sum of energy loss (using the variational form of the governing PDE) and data-driven loss. The labelled high-fidelity datasets for constructing the data-driven loss is simulated using phase field codes developed within the framework of isogeometric analysis (IGA). In this regard, we have used the codes developed in \cite{goswami2020adaptive}, which are available in  \href{https://github.com/somdattagoswami/IGAPack-PhaseField}{https://github.com/somdattagoswami/IGAPack-PhaseField}. However, we note that the proposed approach in this article can also be applied to more general training datasets. For instance, 
one may also consider a V-PINNs algorithm to generate high-fidelity datasets. In our work, we have not considered any experimental data. However, if available, the model can incorporate experimental data to improve the accuracy of the surrogate. To verify the performance of the V-DeepONet based surrogate model, we compare the predicted failure paths to those computed with the high-fidelity fracture mechanics model.

\begin{figure}[htb!]
    \centering
    \subfigure[Tensile loading.]{
    \includegraphics[width = 0.34\textwidth]{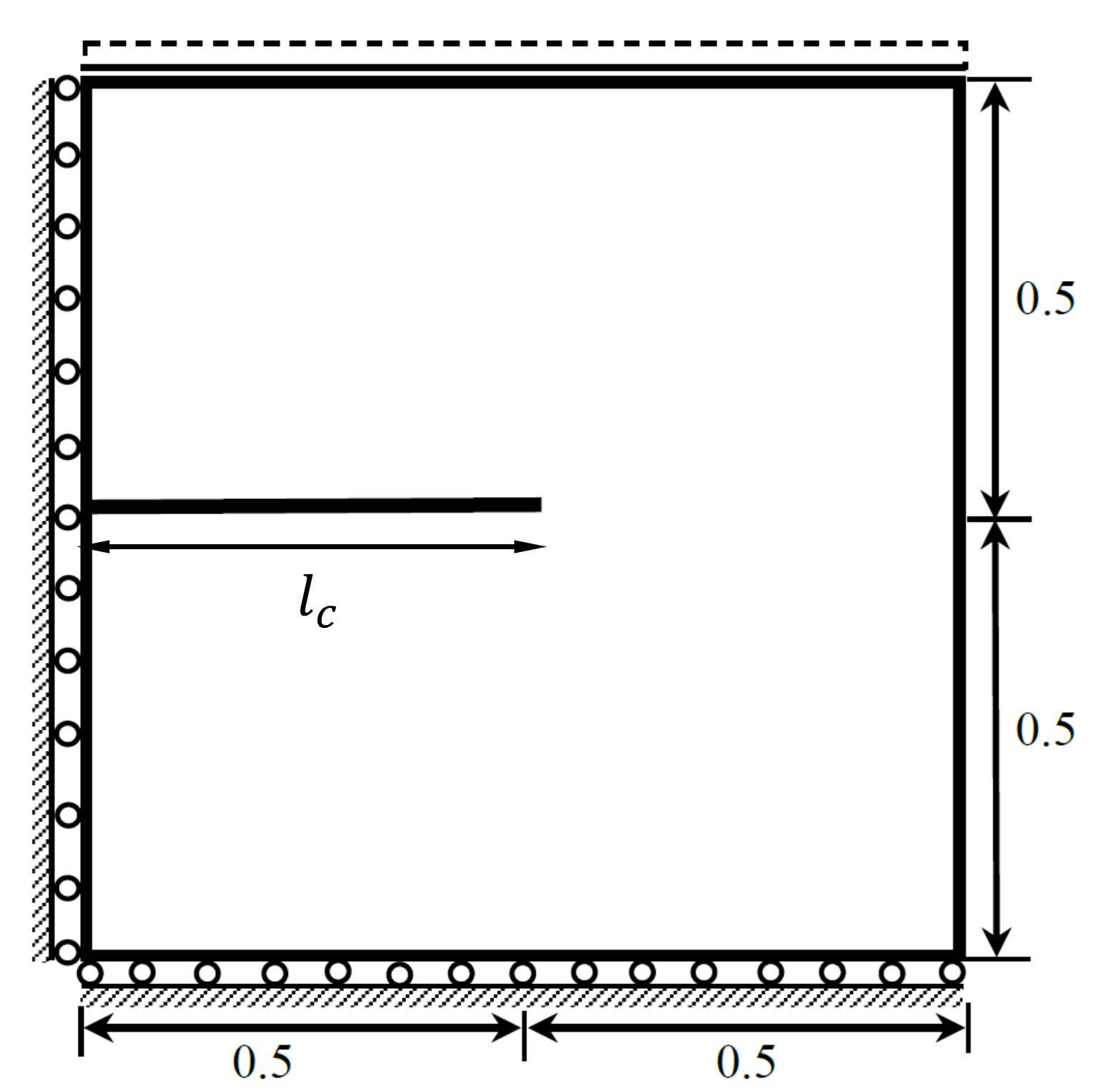}}
    \subfigure[Shear loading.]{
    \includegraphics[width = 0.39\textwidth]{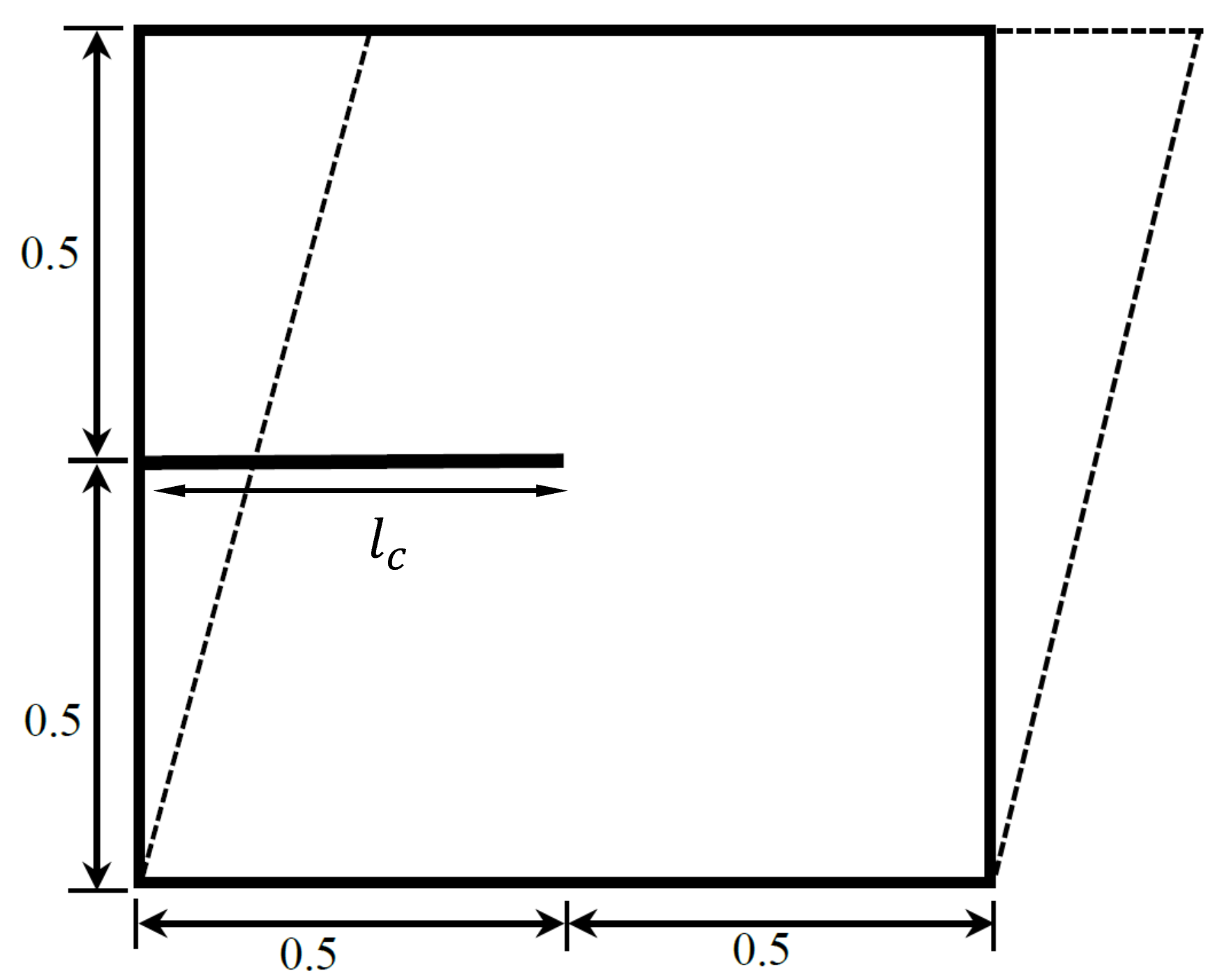}}
    \caption{Geometrical setup and boundary conditions for the single-edge notched plate subjected to different loading conditions \cite{goswami2020adaptive}. All the units are in mm. The crack length, $l_c$ and location shown in the setup diagram is representative of the presence of initial defects.}
    \label{fig:setup}
\end{figure}

\section{Simulation Results}
\label{sec:numericals}

In this section, we explore our method using two benchmark problems from fracture mechanics to demonstrate the effectiveness of the developed surrogate model. In the first problem, we have studied the growth of cracks in a plate under tensile loading (Mode-I failure), while the second one is the growth of cracks under shear loading (Mode-II failure).
For both the problems we have shown the efficiency of the surrogate model to make out-of-distribution predictions. In all examples, the V-DeepONet is trained using the Adam optimizer \cite{kingma2014adam}. The implementation has been carried out using the \texttt{PyTorch} framework \cite{paszke2019pytorch}. Throughout all examples, we will employ the hyperbolic tangent activation function (\texttt{tanh}) and initialize the V-DeepONet parameters using Xavier initialization. Details on the network architecture, such as number of layers, number of neurons in each layer are provided with each example.

Even though the paper focuses on surrogate modeling for predicting brittle failure, we have included a pedagogical problem to show that the proposed V-DeepONet can be used to study general problems with discontinuity. To demonstrate this, we have studied the flow in heterogeneous porous media on a two-dimensional plate with geometric discontinuity. Interested readers may refer to Appendix \ref{app:appendixA}.

\subsection{Brittle fracture in a plate loaded in tension}
\label{subsec:numerical1}

\begin{figure}[htb!]
    \centering
    \subfigure[$\Delta v_1 = 1.4\times 10^{-3}$ mm.]{
    \frame{\includegraphics[trim=40 30 0 20,clip,width = 0.46\textwidth]{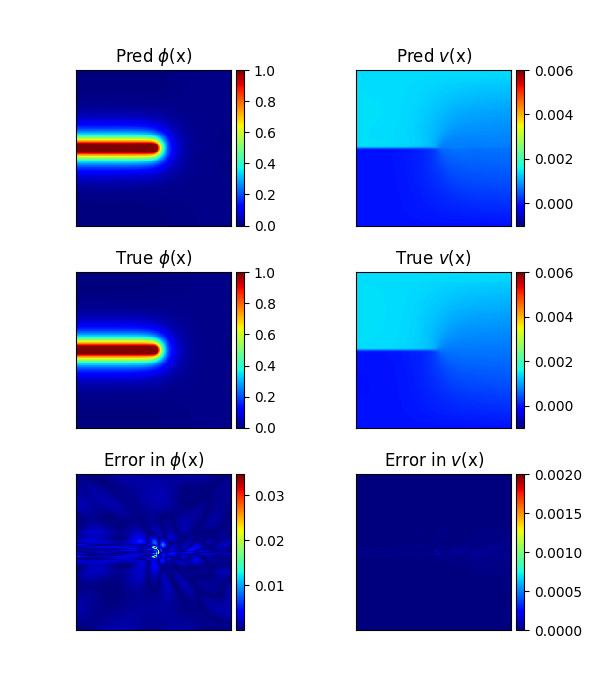}}}
    \subfigure[$\Delta v_5 = 5.0\times 10^{-3}$ mm.]{
    \frame{\includegraphics[trim=40 30 0 20,clip,width = 0.46\textwidth]{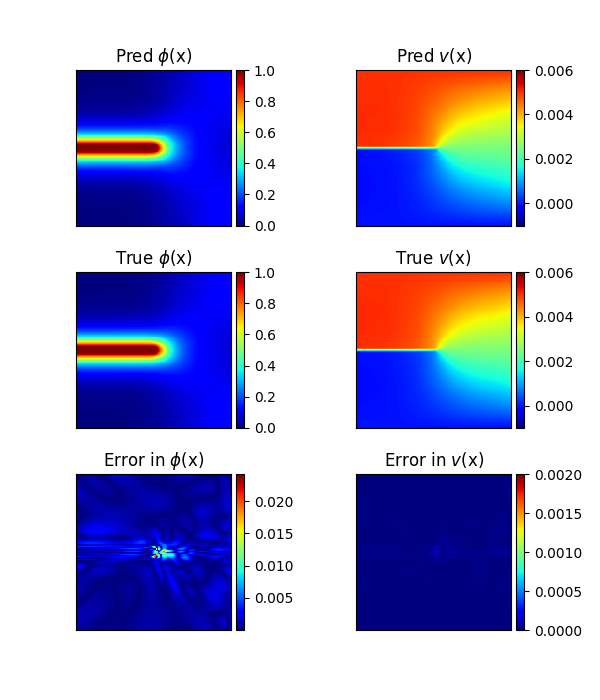}}}
    \subfigure[$\Delta v_6 = 5.6\times 10^{-3}$ mm.]{
    \frame{\includegraphics[trim=40 30 0 20,clip,width = 0.46\textwidth]{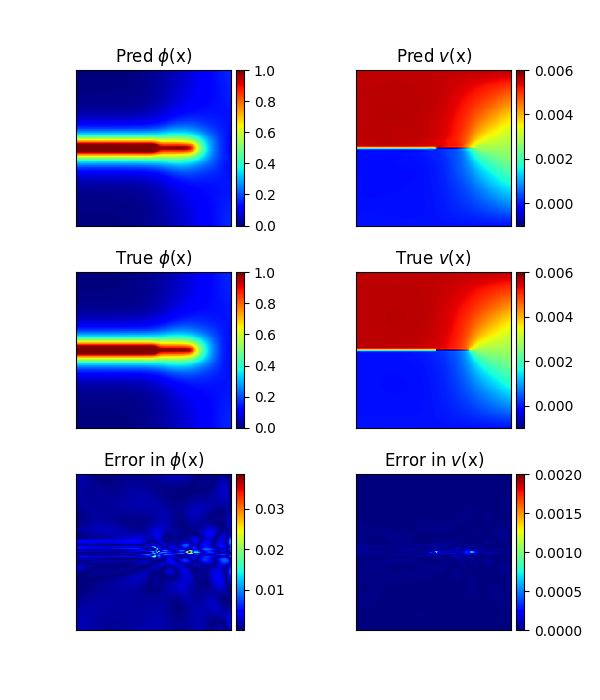}}}
    \subfigure[$\Delta v_7 = 5.8\times 10^{-3}$ mm.]{
    \frame{\includegraphics[trim=40 30 0 20,clip,width = 0.46\textwidth]{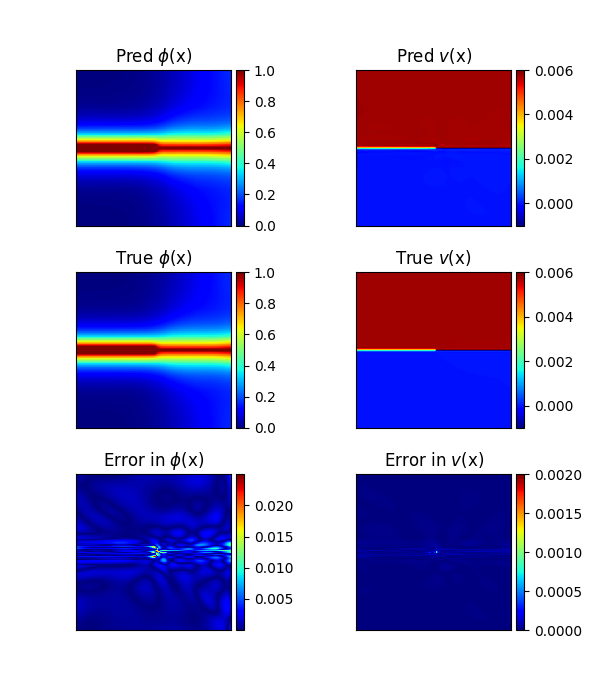}}}  
    \caption{Tensile failure: The V-DeepONet is trained with $6$ values of $l_c \in$ $[0.3,0.55]$ for $7$ displacement steps. The plots for prediction fields with $l_c = 0.5$ mm at certain displacement steps are presented. For each plot, the top row presents the predicted phase field and the displacement along \textit{y}-axis, respectively. The middle row is depicting the ground truth obtained using IGA simulations, while the last row shows the error between the predicted value and the ground truth.}
    \label{fig:tensile_testing}
\end{figure}

\begin{figure}[htb!]
    \centering
    \subfigure[$l_c = 0.5$ mm.]{
   \includegraphics[trim=0 0 0 0,clip,width = \textwidth]{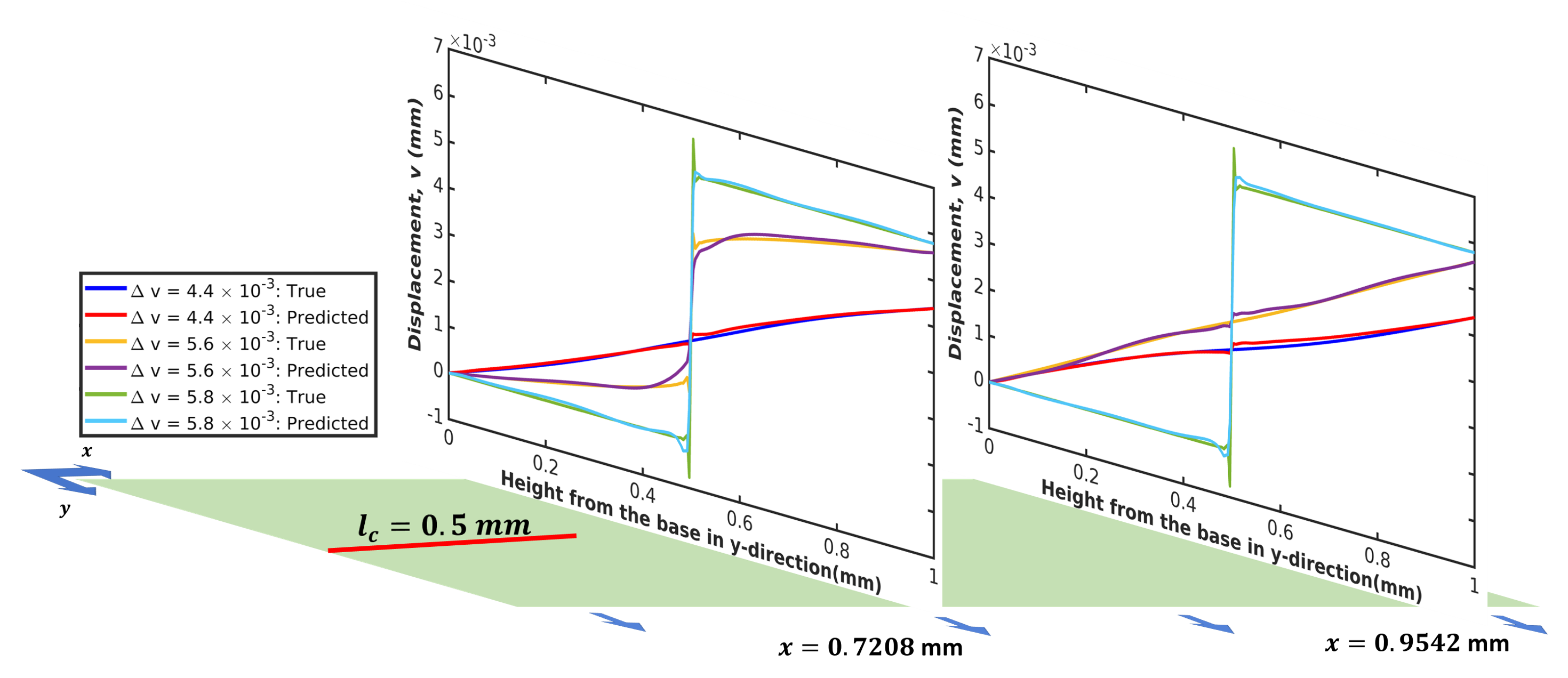}} 
   \subfigure[$l_c = 0.65$ mm]{
   \includegraphics[trim=0 0 0 0,clip,width = \textwidth]{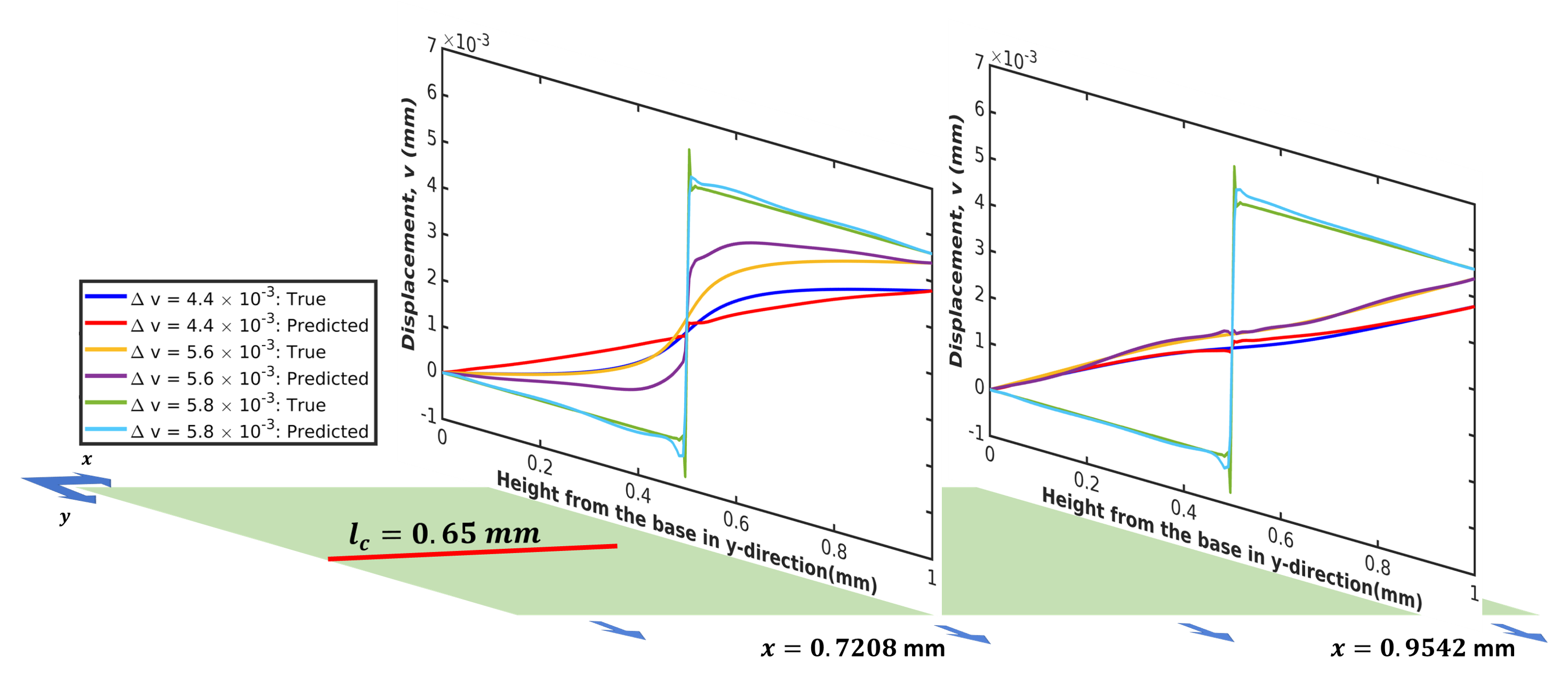}}
   \caption{Tensile failure: The V-DeepONet is trained with $6$ values of $l_c \in$ $[0.3,0.55]$ for $7$ displacement steps. The line plots depict the displacement component in $y-$direction (denoted as $v(x,y)$) at $x_1 = 0.7208$ mm and $x_2 = 0.9542$ mm. Plots in (a) consider $l_c = 0.5$ mm and (b) $l_c = 0.65$ mm (out-of distribution). In each of the plots, $v(x_i,y)$ is plotted as a function of $y$, for $i=1,2$. Different colors of lines represent results from different displacement increment loadings.}
    \label{fig:tensile_testing_line}
\end{figure}

In the first example, we consider a unit square plate with a horizontal crack at the middle height from the left outer edge. The geometrical setup and the boundary conditions of the problem are shown in \autoref{fig:setup}(a). The material properties considered are: $\nu = $ 121.15 kN/mm$^{2}$, $\mu = $ 80.77 kN/mm$^{2}$ and $G_c = 2.7 \times 10^{-3}$ kN/mm. For this problem, we have considered the length scale parameter, $l_0 = 0.0625$ mm. The Dirichlet boundary conditions are:
\begin{equation}
    u(0,y) = v(x,0) = 0, \;\;\; v(x,1)= \Delta v,
\end{equation}
where $u$ and $v$ are the solutions of the elastic field in \textit{x} and \textit{y}-directions, respectively and $\Delta v$ is the applied tensile displacement. In this example, we will train the V-DeepONet model using $s = 6$ initial conditions (by changing the crack lengths, $l_c$, in \autoref{fig:setup}(a)) and consider $r = 7$ displacement steps for obtaining the crack path. The six initial configurations used for training the network are $l_c \in \{0.25, 0.3, 0.35, 0.4, 0.45, 0.55\}$ mm, while the considered displacement steps are $\Delta v \in \{1.4, 2.2, 3.2, 4.4, 5, 5.6, 5.8\}\times 10^{-3}$ mm. To build the surrogate model, $m = 212 $ sensors are chosen close to the cracked region, to distinctly represent each of the function in the input space of the branch net. We compute the initial strain energy, $\bm{\mathcal H}_0^{(i)}$, $i \in \{1,\cdots,s\}$ using \autoref{eq:initial_history_field} at $m$ sensor points. Now, from the high-fidelity data obtained using IGA simulations, we obtain the tensile strain-energy, $\bm{\mathcal H}_j^{(i)}$ for the applied displacements {$\Delta \wb_j$}, where $j\in\{1,\cdots,r-1\}$. The input data for the branch net, $\bm{\mathcal H}(\bm x)$ is prepared in the same way as discussed in \autoref{eq:hist_window}. In this example, we aim to learn the solution operators, $\mathcal G_{\bm{\theta}}^{u}, \mathcal G_{\bm{\theta}}^{v}, \mathcal G_{\bm{\theta}}^{\phi}$ mapping $\bm{\mathcal H}(\bm y)$ to the solutions $u(\bm y), v(\bm y)$, and $\phi(\bm y)$. To this end, we represent the operator by a DeepONet, where both the branch net and the trunk net are 5-layer fully-connected neural networks with 50 neurons per hidden layer and equipped with \texttt{tanh} activations. For each input function in the branch net, samples are provided on $p = 1372$ points in the domain. Once the solution is evaluated at the sampled points, the V-DeepONet outputs for the elastic field are lifted to exactly match the Dirichlet boundary conditions, following:
\begin{equation}
\begin{split}
    \mathcal G_{\bm{\theta}}^{u} &= [x(1-x)]\hat{\mathcal G}_{\bm{\theta}}^{u},\\
    \mathcal G_{\bm{\theta}}^{u} &= [y(y-1)]\hat{\mathcal G}_{\bm{\theta}}^{v} + y\Delta v,
\end{split}
\end{equation}
where $\hat{\mathcal G}_{\bm{\theta}}^{u}$ and $\hat{\mathcal G}_{\bm{\theta}}^{v}$ are obtained from the neural network. The trainable parameters, $\bm{\theta}$ of the V-DeepONet is obtained by minimizing the hybrid loss function in \autoref{eq:DeepOnet_loss}. Now, we test the model to predict the solutions for $l_c = 0.5$ mm. To do this, we compute the initial history function, $\bm{\mathcal H}_0$, analytically using \autoref{eq:initial_history_field} at the sensor locations and also at the sampled points where the solution has to be evaluated. The input to the branch net of the trained model is $\left[\bm{\mathcal H}_0, 0\right]$, while the input to the trunk net are the sampled points, the initial history field at the points sampled in the domain and $\Delta v_1$. The surrogate model predicts the displacement field, $u,v$ and the phase field, $\phi$ corresponding to $\Delta v_1$. In addition, the network predicts the tensile strain energy for the applied displacement, which is denoted as $\bm{\mathcal H}_1$. To predict the solution for an applied displacement of $\Delta v_2$, the input to the branch net is $\left[\bm{\mathcal H}_1, \bm{\mathcal H}_0\right]$, while the input to the trunk net remains the same except for the applied displacement, which is changed to $\Delta v_2$. In a similar way the solution for any crack length at any displacement step can be obtained sequentially. \autoref{fig:tensile_testing} shows the predicted crack path and the displacement component in \textit{y}-axis for some displacement steps, considering $l_c = 0.5$ mm. In \autoref{fig:tensile_testing_line}(a), we have presented the predicted displacement component in $y$-axis against the ground truth at two locations along $x$-axis for three displacement steps. The averaged error of predicted $\phi$ is $0.63\%$.

It is worth investigating that the trained V-DeepONet is capable of yielding accurate predictions for out-of-distribution test data. To illustrate this, we have tested the trained model for predicting the crack path and the displacement fields for $l_c = 0.65$, which is beyond the crack distribution length range $l_c\in[0.25,0.55]$ in our training dataset. In \autoref{fig:tensile_testing_line}(b), we have presented a comparison of the predicted displacement in $y$-axis against the ground truth at $x = 0.7208$ mm and $x = 0.9542$ mm for 3 displacement steps. For the out-of-distribution prediction, an averaged error of $1.85\%$ on $\phi$ is obtained. For interested readers, the predicted solutions obtained using the V-DeepONet based surrogate model, at certain displacement steps, is shown in \autoref{fig:tensile_testing_outofdist} of Appendix \ref{app:appendixC}, where we have compared the predicted results with those simulated using the high fidelity model and the pointwise errors are also plotted.

\subsection{Brittle fracture in a plate loaded in shear}
\label{subsec:numerical2}

\begin{figure}
    \centering
    \subfigure[]{
    \includegraphics[trim=30 30 10 40,clip,width = 0.48\textwidth]{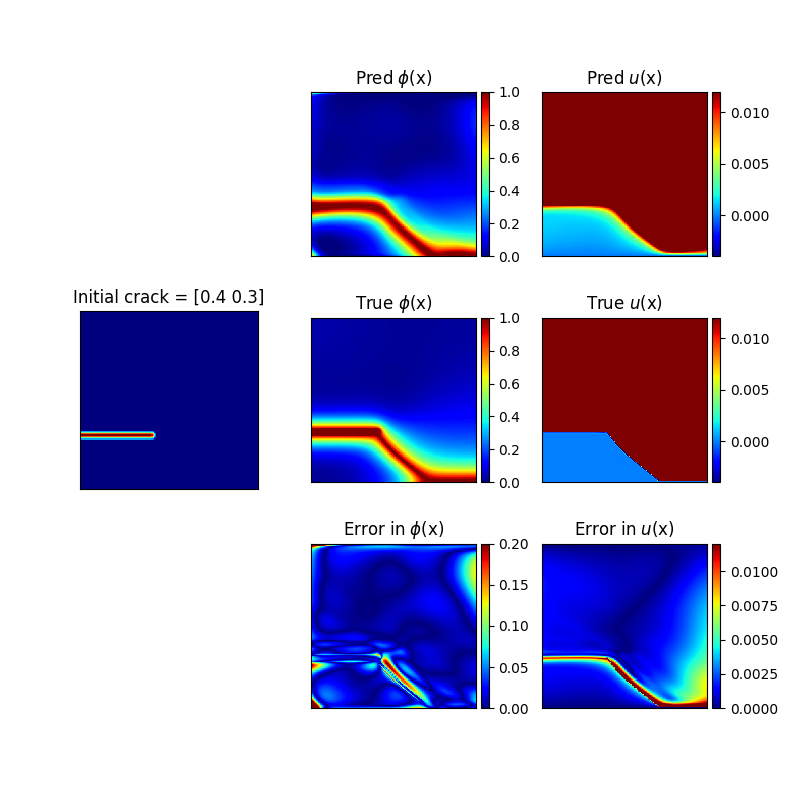}}
    \subfigure[]{
    \includegraphics[trim=30 30 10 40,clip,width = 0.48\textwidth]{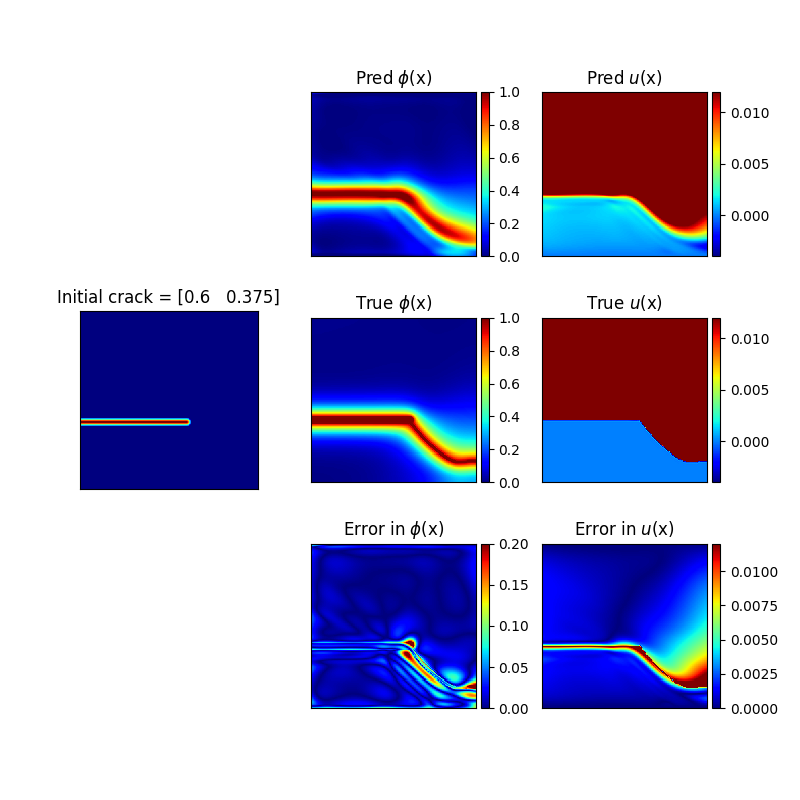}}
    \caption{Shear failure: The V-DeepONet is trained with $85$ samples with crack tips located throughout the domain to predict the final damage path. The plots for two test cases are shown where the crack tips are located at: (a) $(0.4,0.3)$, (b) $(0.6,0.375)$. For each plot, the initial configuration is shown on the left. The top row presents the predicted phase field, displacement along \textit{x}-axis, and displacement along \textit{y}-axis, respectively. The middle row is depicting the ground truth obtained using the simulations in IGA, while the last row shows the error.}
    \label{fig:shear_Ytest_final}
\end{figure}

In this example, we investigate the same square plate as stated in \autoref{subsec:numerical1} for a pure shear loading mode. The geometrical setup and the boundary conditions of the problem are shown in \autoref{fig:setup}(b). We consider the same material parameters as used in \autoref{subsec:numerical1}. The Dirichlet boundary conditions are:
\begin{equation}
    u(x,0) = v(x,0) = 0, \;\;\; u(x,1)= \Delta u,
\end{equation}
where $u$ and $v$ are the solutions of the elastic field in \textit{x} and \textit{y}-axis, respectively and $\Delta u$ is the applied shear displacement on the top edge of the plate. In this example, we have trained the V-DeepONet using $n=85$ initial crack locations, with aiming to predict the final crack path for any initial crack location in the domain. The training sample consists of crack lengths, $l_c \in [0.3, 0.65]$ and the height of the crack varied between $[0.2,0.675]$. \autoref{fig:shear_Ytest_final} presents the predicted solution of the elastic field and the phase field for two samples. The predicted $\phi$ has a relative error of $0.67\%$.

\begin{figure}
    \centering
    \subfigure[]{
    \includegraphics[trim=40 20 0 5,clip,width = 0.75\textwidth]{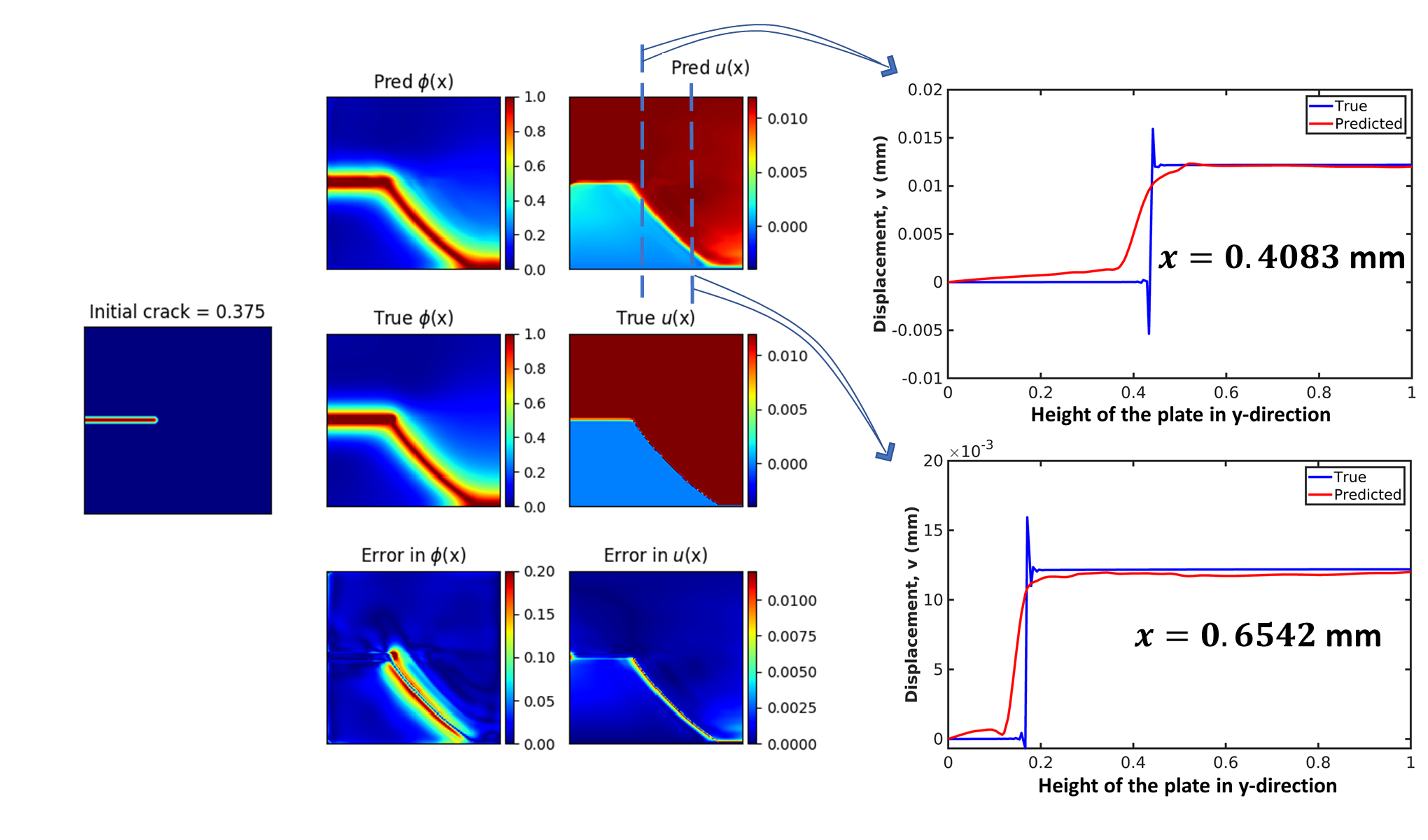}}
    \subfigure[]{
    \includegraphics[trim=40 20 0 5,clip,width = 0.75\textwidth]{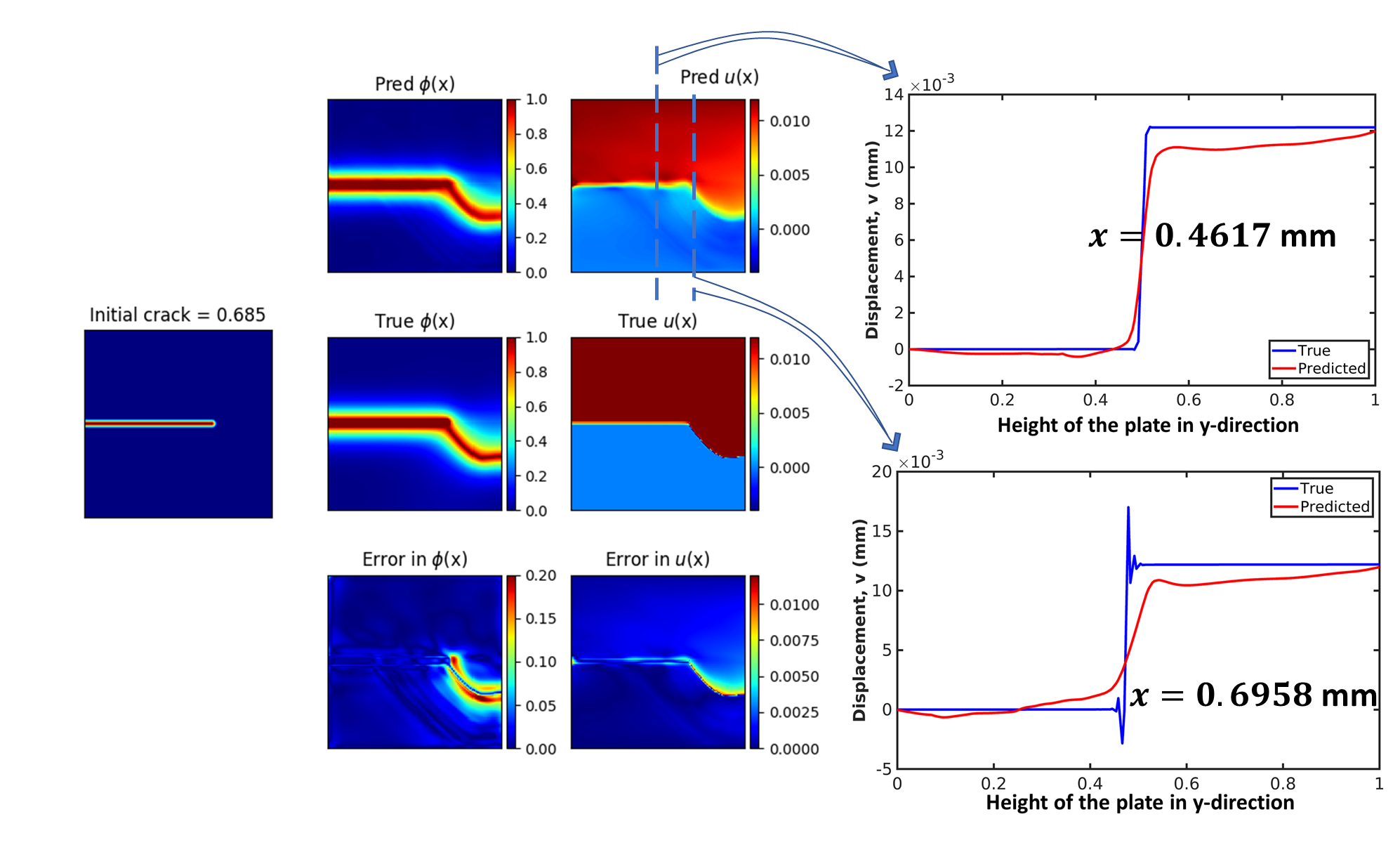}}
    \caption{Shear failure: The V-DeepONet is trained with $11$ crack lengths to predict the final damage path for any crack length, when the height of the crack is fixed at the centre of the left edge. The plots are for (a) $l_c = 0.375$ mm and (b) $l_c = 0.685$ mm, where $\Delta u = 0.220$ mm. For each plot, the predicted displacement in $x$-direction is plotted for two locations along the $x$-axis and is compared with ground truth to show the accuracy of the prediction.}
    \label{fig:shear_Xtest_final}
\end{figure}

\begin{figure}
    \centering
    \subfigure[]{
    \includegraphics[trim=40 20 0 10,clip,width = 0.75\textwidth]{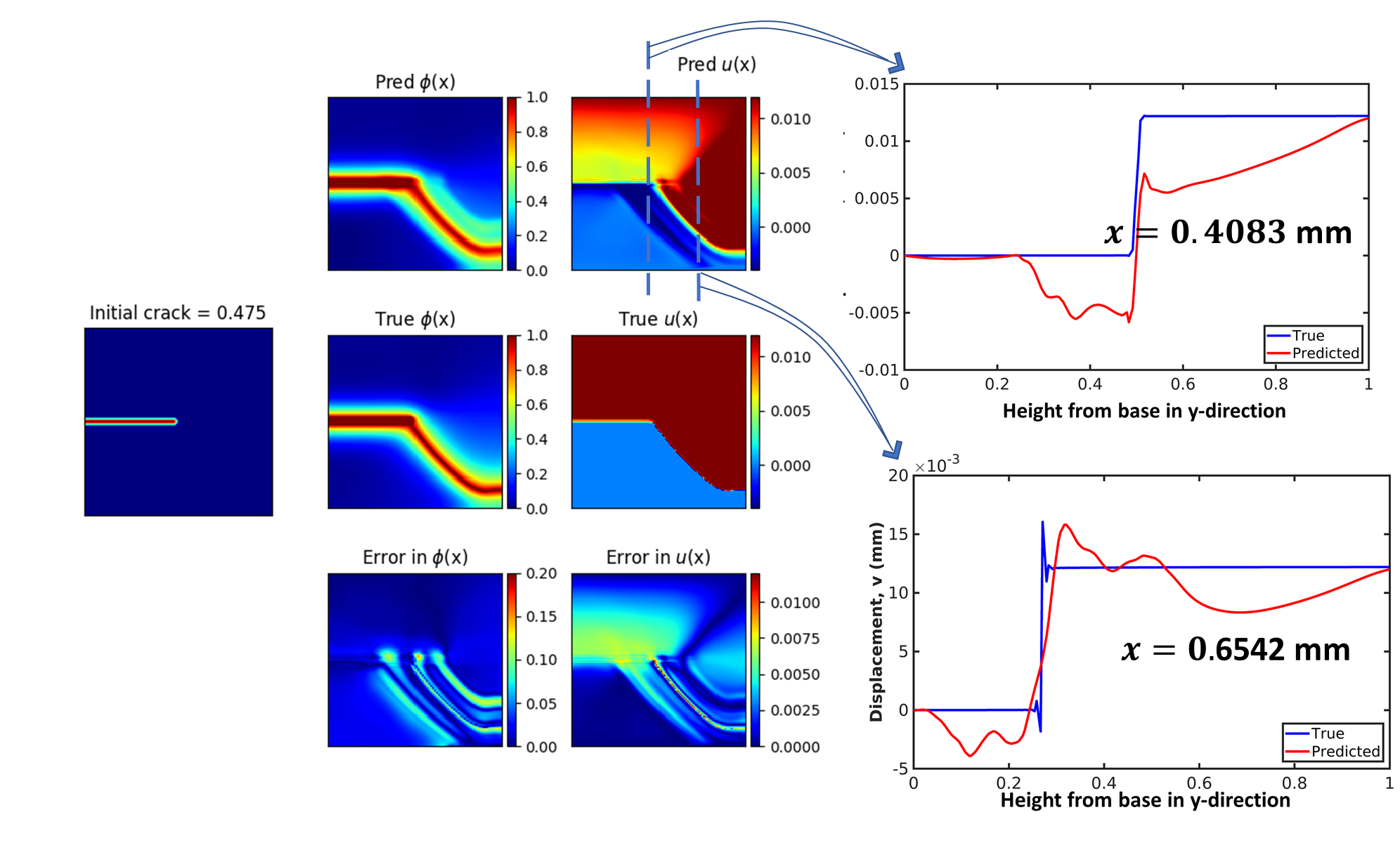}}
    \subfigure[]{
    \includegraphics[trim=40 10 0 10,clip,width = 0.75\textwidth]{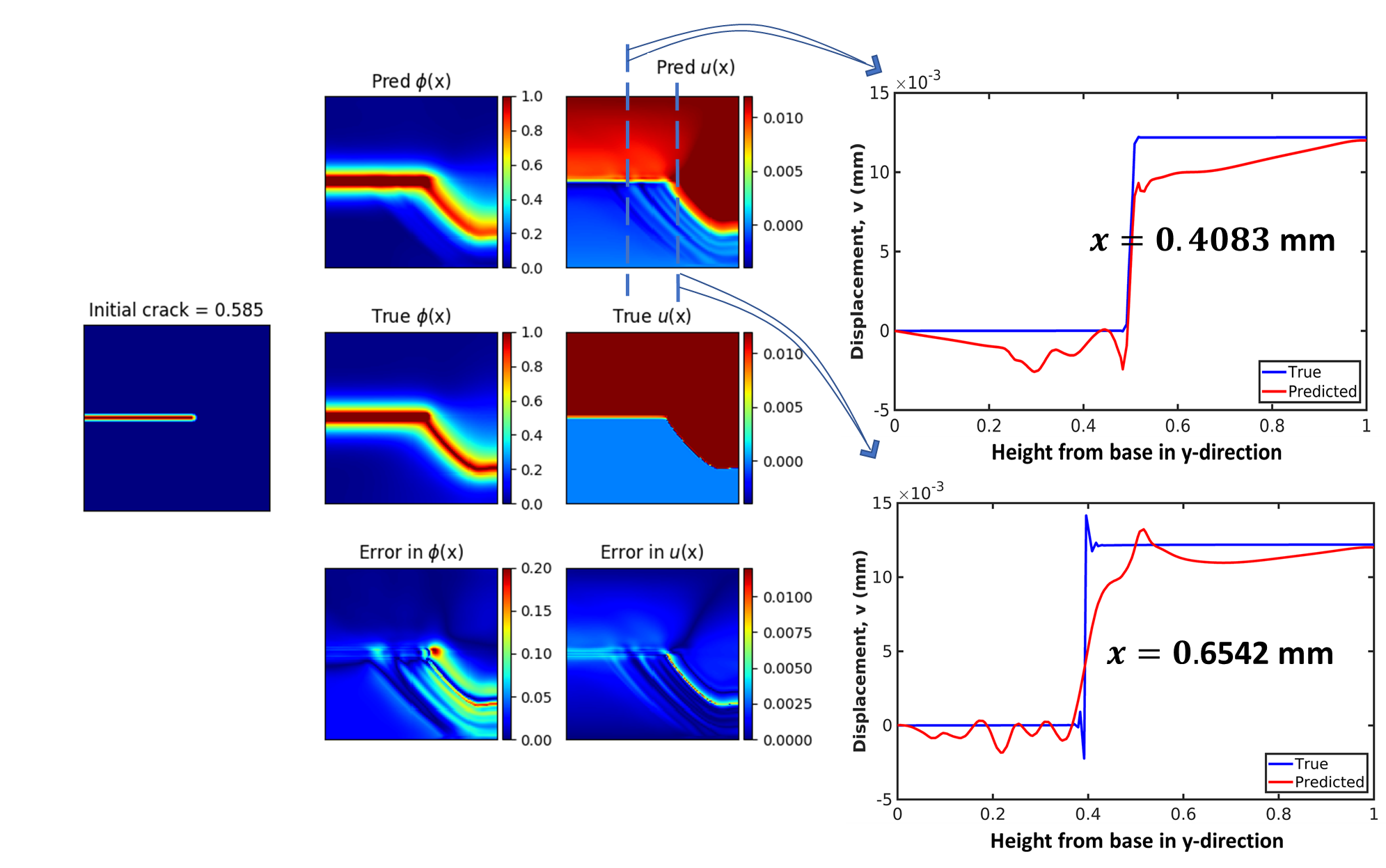}}
    \caption{Shear failure: the data-driven DeepONet (original) is trained with $43$ crack lengths to predict the final damage path for any crack length, when the height of the crack is fixed at the centre of the left edge. The plots are for (a) $l_c = 0.475$ mm and (b) $l_c = 0.585$ mm, where $\Delta u = 0.220$ mm. For each plot, the predicted displacement in $x$-direction is plotted for two locations along the $x$-axis and is compared with ground truth to show the accuracy of the prediction.}
    \label{fig:shear_Xtest_final_DD}
\end{figure}

Now, we have carried out multiple comprehensive studies to depict the versatility of the proposed surrogate model. 
In the first experiment, we compare the accuracy and efficiency of V-DeepONet with solely data driven DeepONet\cite{lu2021learning}. To that end, we consider changing the crack length, $l_c$, while fixing the height of the crack at the middle of the left outer edge, and predict the final damage path. To train the surrogate model we have used $n = 11$ samples with initial crack lengths, $l_c \in [0.2, 0.7]$ in steps of $0.05$. For this experiment, we have used the surrogate model as discussed in \autoref{subsec:surrogate2} and have considered $m = 934$ sensors, closely placed around the region of high stresses. The initial strain energy, $\bm{\mathcal H}(\bm y_j)$, where $j\in\{1,\cdots,m\}$, is computed using \autoref{eq:initial_history_field} for all the $l_c$, which is used as input to the branch net. For the trunk net, $p = 6024$ points are sampled in the domain. The input to the trunk net is a tensor of dimension $(n\times p,3)$, where the first two columns contain the spatial locations of the sampled points, while the third column corresponds to the initial strain energy computed at the sampled points for each of the $n$ cases. In this experiment, we aim to learn the solution operators, $\mathcal G_{\bm{\theta}}^{u}, \mathcal G_{\bm{\theta}}^{u}, \mathcal G_{\bm{\theta}}^{\phi}$ so as to map $\bm{\mathcal H}(\bm y)$ to their solutions $u(\bm y), v(\bm y)$, and $\phi(\bm y)$ for predicting the final damage path considering $\Delta u = 1.2 \times 10^{-2}$mm. To this end, we represent the operators by a V-DeepONet, where both the branch net and the trunk net are 4-layers fully-connected neural networks with $[100, 50, 50, 50]$ neurons, respectively. Once the solution is evaluated at the sampled points, the outputs for the elastic field are modified to exactly satisfy the Dirichlet boundary conditions, as:
\begin{equation}\label{eq:shear_boundary_auto}
\begin{split}
    \mathcal G_{\bm{\theta}}^{u} &= [y(1-y)]\hat{\mathcal G}_{\bm{\theta}}^{u} + y \Delta u,\\
    \mathcal G_{\bm{\theta}}^{u} &= [y(y-1)]\times [x(x-1)] \hat{\mathcal G}_{\bm{\theta}}^{v}, 
\end{split}
\end{equation}
where $\hat{\mathcal G}_{\bm{\theta}}^{u}$ and $\hat{\mathcal G}_{\bm{\theta}}^{v}$ are obtained from the DeepONet. The trainable parameters of the V-DeepONet are obtained by minimizing the hybrid loss function in \autoref{eq:DeepOnet_loss}. The trained surrogate model is used to predict the final crack path and solutions of the elastic field for $l_c = 0.375$ mm and $l_c = 0.685$ mm. The predicted plots are  shown in \autoref{fig:shear_Xtest_final}.
The training trajectory of the V-DeepONet using $11$ training samples is shown in \autoref{fig:shear_X_error} in Appendix \ref{app:appendixC}. In the plot, the training loss depicts the hybrid loss given by \autoref{eq:DeepOnet_loss}. The conventional DeepONet is trained with the same $11$ samples, keeping the network architecture of the branch net and the trunk net exactly the same. A prediction error of $26.2\%$ is reported when trained with $11$ training samples. To improve the accuracy, the training samples are increased to $22$, but the model is unable to capture the mode-II failure. Lastly, the number of training samples is increased to $43$ and a relative mean error of $3.12\%$ is reported for $\phi$. \autoref{fig:shear_Xtest_final_DD} presents the plots of the predicted solutions for $l_c = 0.475$ mm and  $l_c = 0.585$ mm when the conventional DeepONet is trained with $43$ samples. The predicted results for the data driven DeepONet depict that it is unable to capture the crack diffusion phenomenon and also it cannot generalise complex fracture phenomenon with limited data-sets.

\begin{figure}
    \centering
   \includegraphics[width = \textwidth]{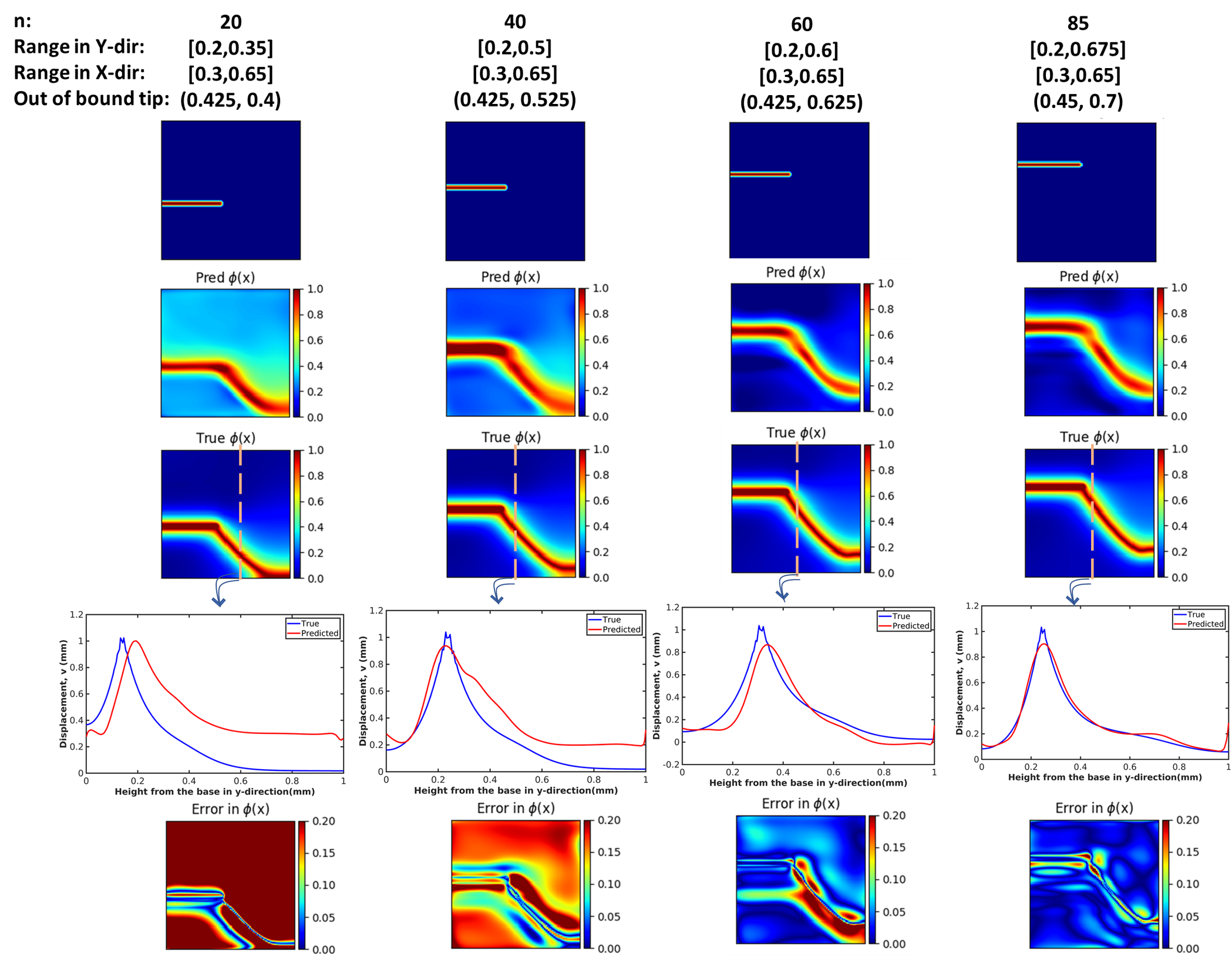}
    \caption{Shear failure: The V-DeepONet is trained with $n=20,$ $40,$ $60,$ $85$ samples (from left to right). For each $n$, a corresponding out-of distribution prediction is made. It is observed that the V-DeepONet can generalize the solution with just $85$ training samples. A cross-section taken at $x=0.6048$ is taken to show the precdicted $\phi$ against the ground truth. The relative mean error is reported as $26.44\%$, $13.49\%$, $7.49\%$ and $3.17\%$ (from left to right).}
    \label{fig:shear_Ytest_OutofD_phi}
\end{figure}

\begin{figure}
    \centering
    \subfigure[]{
    \fbox{\includegraphics[width = 0.4\textwidth]{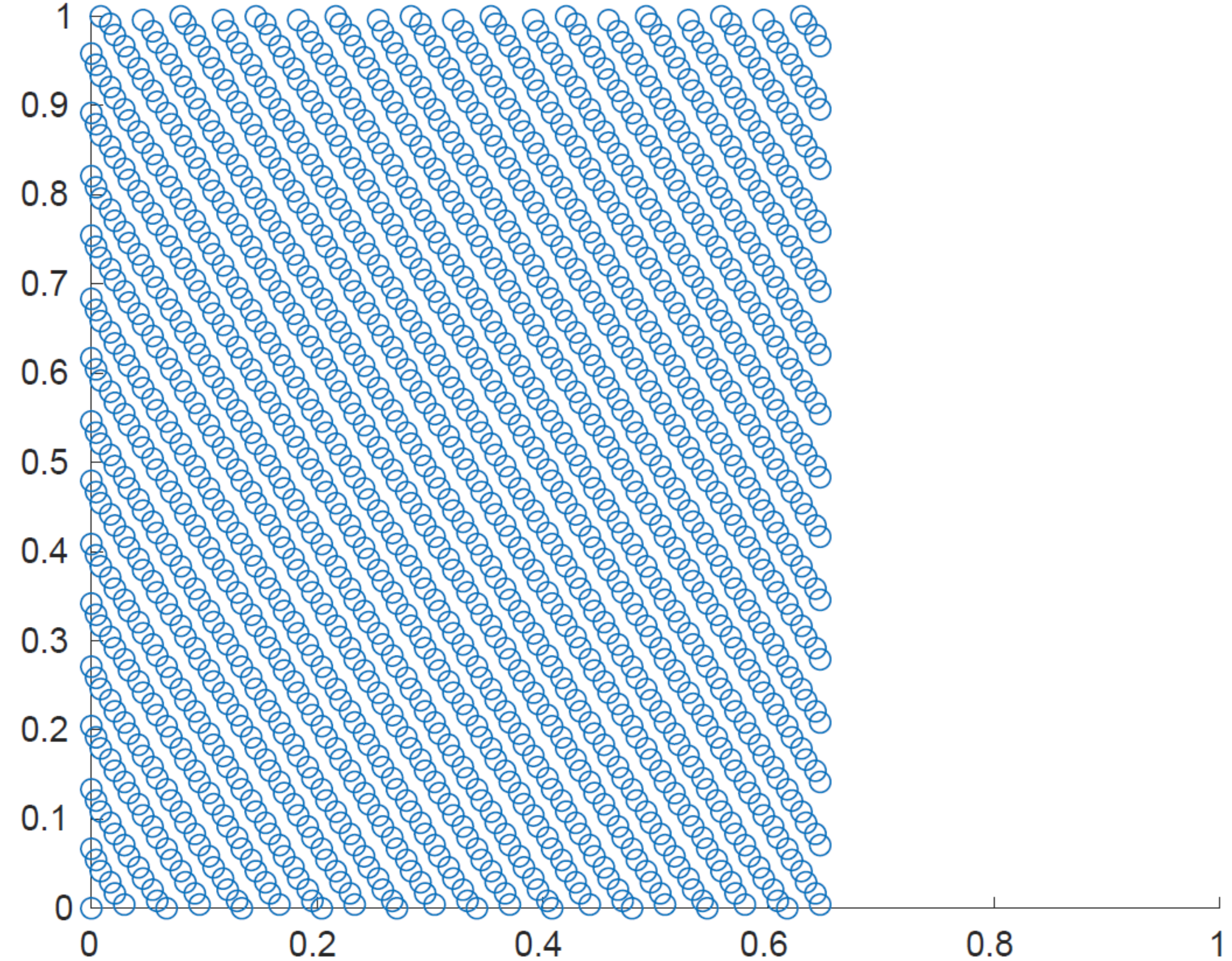}}}
    \subfigure[]{
    \fbox{\includegraphics[trim=0 17 0 0,clip,width = 0.43\textwidth]{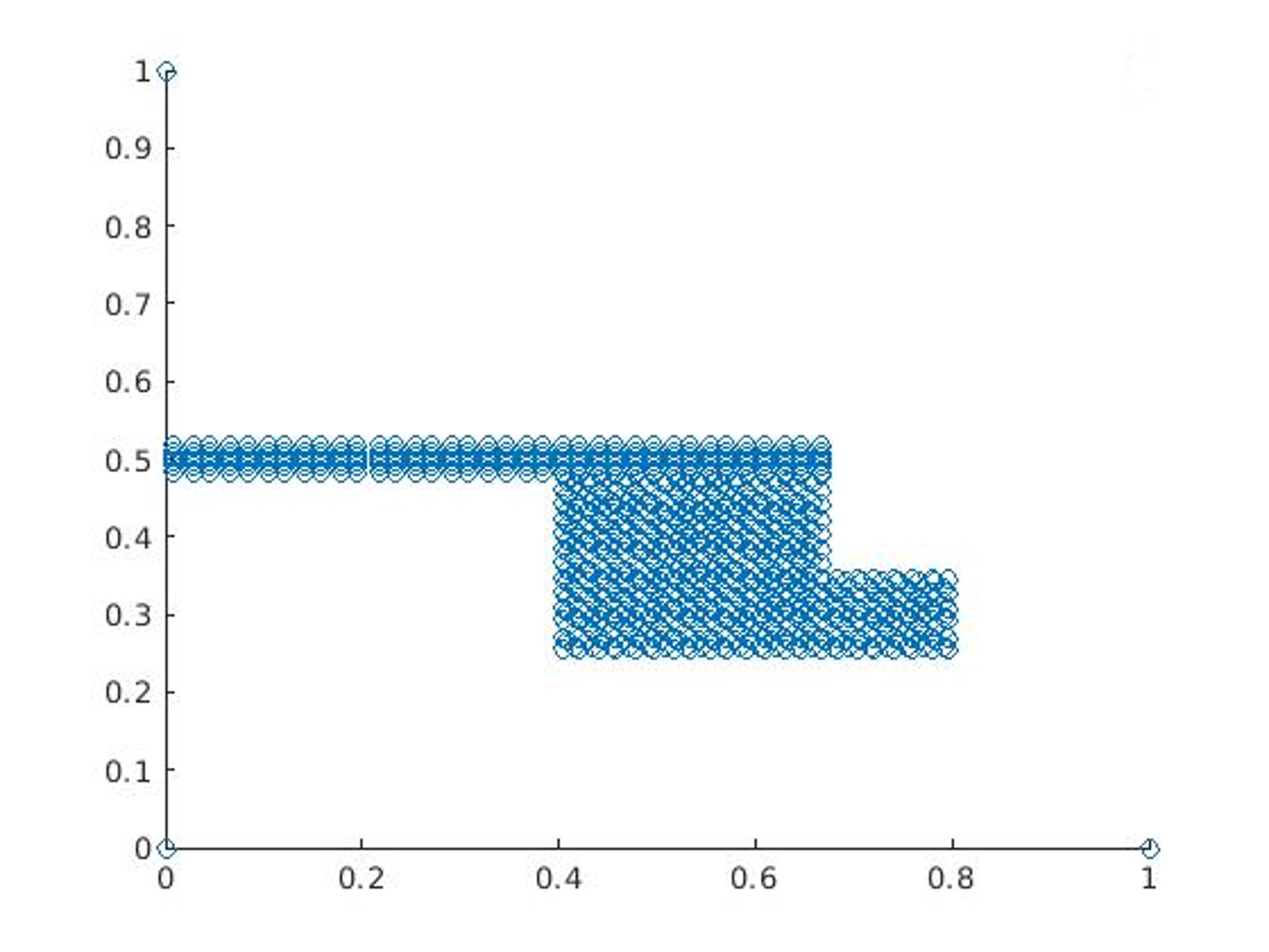}}}
    \caption{Shear failure: (a) The $m=1547$ sensor points are shown using blue circles. The sensors are chosen to represent the input functions discretely, so that network approximations can be applied. In this example, the training samples are chosen such that the initial crack can be placed anywhere between $[0,1]$ in the \textit{y}-axis. However, the initial crack lengths are restricted between $[0.2,0.65]$. So, the sensor locations are between $[0,0.65]$ along the horizontal axis and between $[0,1]$ along the vertical axis. (b)The $m= 854$ sensor points chosen to represent the input functions discretely as the crack grows.}
    \label{fig:shear_XY_error}
\end{figure}

In the next experiment, we change the location of the crack vertically and at the same time we change the length of the horizontal crack. The aim of this experiment is to find the final damage path for any given location of the crack in the domain and any crack length and also to test the capability of the model to make an out-of-distribution prediction. To start with, $n=20$ initial configurations have been considered to train the V-DeepONet, with crack lengths varying between $[0.3, 0.65]$, while the height of the crack is between $[0.2,0.35]$. The trained model is used to predict the crack path for an initial crack tip located beyond the training data range, at $(0.425, 0.4)$. The predicted $\phi$ has an error $26.44\%$. To improve the generalization of the V-DeepONet, we add 20 more samples to the training set, hence $n=40$. The length of crack varies over the same range as the previous case, however the height of crack varies between $[0.2, 0.5]$. The trained V-DeepONet is used to predict the final crack path for a crack tip located at $(0.425,0.525)$, which is again beyond the range of the training samples. The predicted $\phi$ is reported to have $13.49\%$ error. To the 40 training samples, another 20 training samples are added, making $n=60$. In this case the height of the crack varies in the range $[0.2,0.6]$. The crack path is predicted for a crack tip located at $(0.425,0.625)$, which is again beyond the training range. The prediction accuracy of the model is improved, and an error of $7.49\%$ is reported in this case. In the last case, 25 training samples are added, hence $n=85$. The crack height varies between $[0.2,0.675]$ and the crack length varies between $[0.3,0.65]$ in the 85 training samples. The V-DeepONet based surrogate model is used to predict the crack path for a crack tipped at $(0.4,0.7)$, an out-of-bound sample. The predicted $\phi$ has an error of $3.17\%$. The prediction plots of $\phi$ against the ground truth at a cross-section located at $x = 0.6048$ mm and also error plot over the whole domain is presented in \autoref{fig:shear_Ytest_OutofD_phi}. In this experiment, $m = 1547$ sensors are considered, which is pictorially shown in \autoref{fig:shear_XY_error}(a). To construct the hybrid loss function, $p = 6024$ points are sampled in the domain. The solution is approximated with a V-DeepONet, where the branch and trunk networks are two separate 4-layer fully-connected neural networks with $[100, 100, 50,50]$ neurons, respectively. For plots presenting the prediction of the elastic field solutions, readers may refer to \autoref{fig:shear_Ytest_OutofD} in Appendix \ref{app:appendixC}. The out-of-distribution prediction accuracy increases with the number of samples since an over-parametrized neural network can generalize better. The accuracy of the model in making an out-of-bound prediction with just $85$ training samples, is attributed to the hybrid loss function which integrates physics with the data loss. In \autoref{fig:shear_Ytest_final}, we present the predicted final crack path and the solutions for the elastic field for two crack tips located at $(0.4,0.3)$ and $(0.6,0.375)$, considering $\Delta u = 0.0120$ mm, when the V-DeepONet model is trained with $85$ samples. 

In the last experiment we train the V-DeepONet to obtain the crack location at various displacement steps for different initial conditions. For this experiment, we consider a $s = 10$ initial conditions and $r=3$ displacement steps, $\Delta u = \{1.04, 1.08, 1.14\}\times 10^{-2}$ mm, where we have fixed the height of the crack at the center of the left edge and varied the initial crack length in a range such that $l_c \in [0.4,0.55]$. The selection of appropriate $m = 824$ sensors points is essential in this experiment. In \autoref{fig:shear_XY_error}(b), we show the location of sensor points chosen for this experiment. It is essential that the sensor points are well placed to accurately capture the strain energy for all the displacement steps for all the initial conditions, and must be distinguishable from one another. We compute the initial strain energy, $\bm{\mathcal H}_0^{(i)}$, $i \in \{1,\cdots,r\}$ using \autoref{eq:initial_history_field} at $m$ sensor points. From the high-fidelity data simulated using IGA, we obtain the tensile strain-energy, $\bm{\mathcal H}_j^{(i)}$ for the applied displacements $\Delta u_j$, where $j\in\{1,\cdots,s\}$. The input data for the branch net, $\bm{\mathcal H}(\bm y)$ is prepared in the same way as discussed in \autoref{eq:hist_window}. In this example, we aim to learn the solution operators, $\mathcal G_{\bm{\theta}}^{u}, \mathcal G_{\bm{\theta}}^{v}, \mathcal G_{\bm{\theta}}^{\phi}$ from $\bm{\mathcal H}(\bm y)$ to approximate the solution of $u(\bm y), v(\bm y)$, and $\phi(\bm y)$. To that end, we represent the operators by a V-DeepONet, where both the branch net and the trunk net are 4-layer fully-connected neural networks with 100 neurons per hidden layer. For each input function in the branch net, $p = 6024$ points are sampled in the domain. The ground truth is obtained from the high fidelity solver. Once the solution is evaluated at the sampled points, the V-DeepONet outputs for the elastic field are modified to exactly satisfy the Dirichlet boundary conditions using \autoref{eq:shear_boundary_auto}. The hybrid loss function is then constructed as the sum of the data loss and the total energy. The trainable parameter of the V-DeepONet is then computed by minimizing the hybrid loss using the Adam optimizer \cite{kingma2014adam}.

\begin{figure}
    \centering
    \subfigure[]{
    \includegraphics[trim=10 40 10 40,clip,width = 0.47\textwidth]{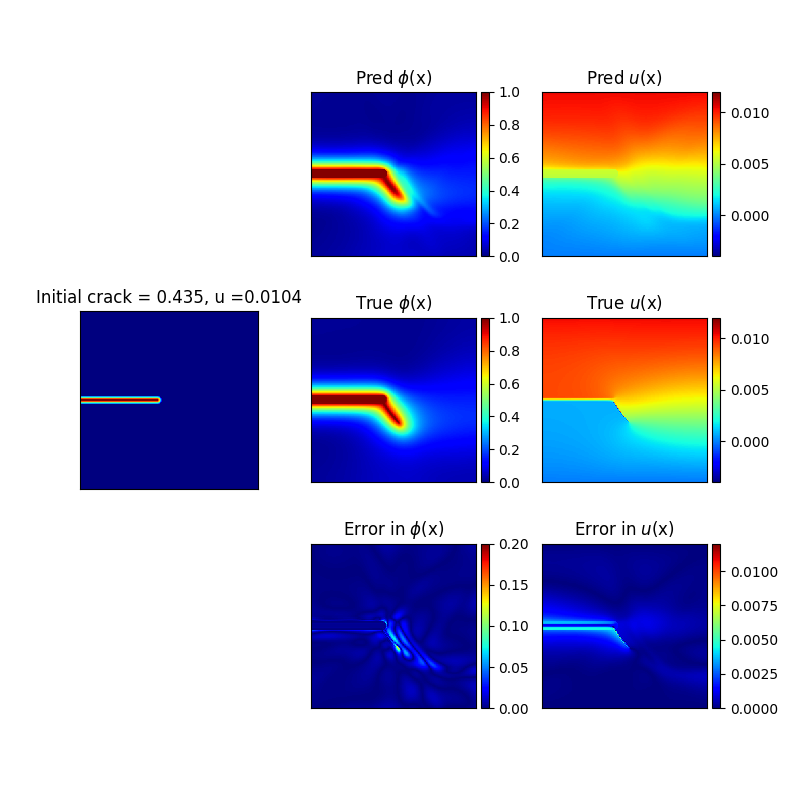}}
    \subfigure[]{
    \includegraphics[trim=10 30 10 40,clip,width = 0.47\textwidth]{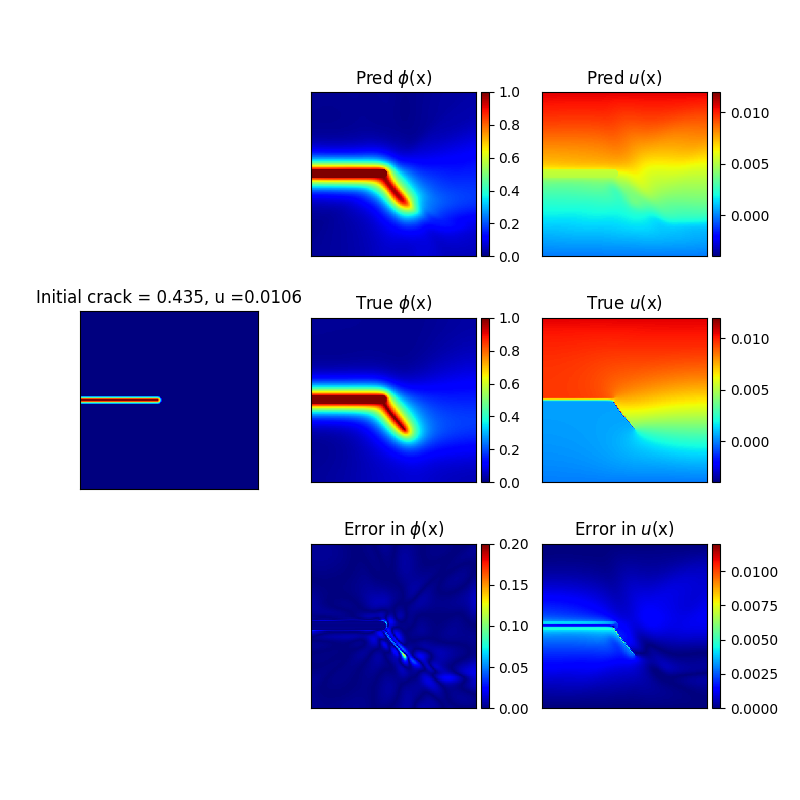}}
    \subfigure[]{
    \includegraphics[trim=10 30 10 40,clip,width = 0.47\textwidth]{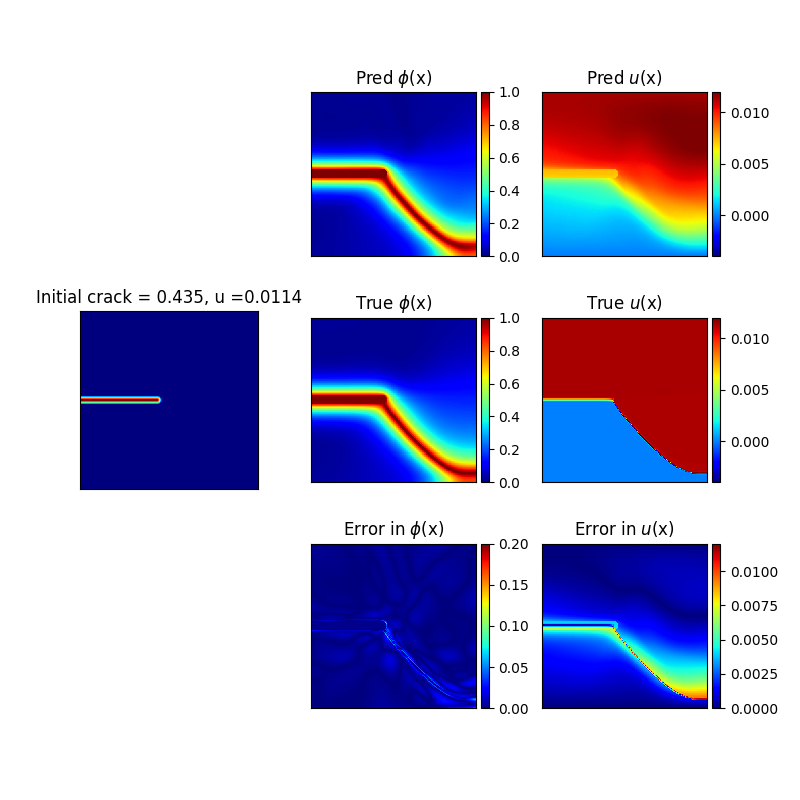}}
    \caption{Shear failure: V-DeepONet is trained with $10$ initial crack configuration for $3$ displacement steps. The plots for a testing sample with initial crack length $=0.435$ mm at 3 displacement steps are shown. (a) $\Delta u=0.0104$ mm, (b) $\Delta u=0.0106$ mm, (c) $\Delta u=0.0114$ mm. For each plot, the initial configuration is shown on the left with the applied displacement at that step. The top row presents the predicted phase field and the displacement along \textit{x}-axis. The middle row is depicting the ground truth obtained using the simulations in IGA, while the last row shows the error.}
    \label{fig:shearpath_testing}
\end{figure}

Now, we use the trained model to predict the solutions of phase field and the displacement fields at all the displacement steps for $l_c = 0.435$ mm. For predicting the crack position at $\Delta u = 0.0104$ mm, the input to the branch net is the strain energy analytically obtained using \autoref{eq:initial_history_field} at $854$ sensor locations padded with $854$ zeros, indicating that there is no history behind. The solution for the field variables are obtained and then strain energy is computed at $\Delta u = 0.0104$ mm. To predict the displacement field at $\Delta u = 0.0106$ mm (different from the trained displacements), the input to the branch net is the current strain energy and the initial strain energy. The surrogate model predicts the displacement field, $u,v$ and the phase field, $\phi$ corresponding to all the displacement steps sequentially. The predicted plots for $l_c = 0.435$ mm are shown in \autoref{fig:shearpath_testing}.

\section{Summary and discussion}
\label{sec:conclusion}

We have proposed neural networks to address the limitations of numerical methods that can simulate a single set of I/BCs and domain geometry at a time. The model developed in the framework of V-DeepONet provides an efficient surrogate of high-fidelity simulations to estimate key quantities of interest for brittle fracture such as failure paths, failure zones, and damage along failure. Once the V-DeepONet is trained, the model can predict failure paths and the corresponding displacements for any location of the crack tip and at any applied displacement at a fraction of a second. The salient features of the proposed model are:
\begin{enumerate}
    \item The model considers a hybrid loss function for training, which is the weighted sum of physics driven loss and data loss. This makes it ideal for extrapolation tasks. 
    \item The physics loss is derived from the variational form of the governing PDE, which makes it a very good choice for problems with discontinuous solutions.
    \item Scaling to larger and more complex problems is often computationally prohibitive with computational tools such as molecular dynamics. This approach is independent of the dimensionality.
    \item An extrapolation accuracy of $3.17\%$ is reported for a Mode-II failure, which is often a challenging problem to solve even for classical numerical methods.
\end{enumerate}
The proposed methodology is ideal for usage in domains such as  reliability analysis, uncertainty quantification and design optimization.

Despite the excellent performance of the proposed surrogate model, it is important to note that this approach is handling fracture problems, which is very sensitive to minor fluctuations and hence, has certain limitations. First and foremost is the data-scaling issue. The input to the feed-forward network is energy, which is in the range $\left[0,10^{4}\right]$ for a single step. Scaling the input tensile energy is essential but it has to be scaled judiciously since the input functions at the sensor locators have to be unique and distinguishable. 
Secondly, the weights of the loss-terms $\lambda_1$ and $\lambda_2$ have to be modulated manually to strike a balance between the data-driven loss and the energy loss. 
Finally, the outputs of the V-DeepONet are the elastic field, which is of the order $O(10^{-3})$, and the phase field, which is of the order $O(1)$. Hence, a scaling of the individual outputs of the V-DeepONet is important before computing the energy and the data-driven loss.
In future works, we will address some of these issues and adopt an adaptive approach to choose the weights.

\section*{Acknowledgement}
\label{sec:acknowledgement}
This work was funded by the DOE PhILMs project (no. DE- SC0019453), OSD/AFOSR MURI grant FA9550-20-1-0358, and National Institutes of Health grant (U01 HL142518). Y. Yu was supported by the National Science Foundation under award DMS 1753031. The authors thank Dr. Xuhui Meng for helpful discussion and his Darcy's problem simulation code. S. Goswami thanks Dr. Khemraj Shukla for his support on setting on high-performance computing environment.  

\begin{appendices}

\section{Algorithm for constructing the proposed unified surrogate model.}
\label{app:appendixB}

\noindent{Pre-requirement of data from the high-fidelity solver:}
\begin{enumerate}
    \item Sample the domain for $p$ points, $\{\bm y_i\}_{i=1}^p$, where the V-DeepONet will be evaluated.
    \item For each of the $n$ initial defect locations, obtain the responses, $u(\bm y), v(\bm y),\phi(\bm y)$ for selected $r$ displacement steps, $\{\Delta \wb_i\}_{i=1}^r$.
    \item Compute the tensile strain energy, $\bm{\mathcal H}_i$ for $\{\Delta  \wb_i\}_{i=1}^{(r-1)}$, corresponding to $n$ cases.
\end{enumerate}
We put forth the algorithm for constructing the surrogate model to predict the crack path for brittle fracture in \autoref{alg:V-deepONet_surr}.
\begin{algorithm}[!htbp]
\caption{V-DeepONet-based surrogate model for predicting brittle fracture.}\label{alg:V-deepONet_surr}
\textbf{Inputs:} The location of $m$ sensors, $\mathcal X_s = \{\bm x_1, \bm x_2 \ldots, \bm x_m\}$.\\
Compute the initial history function, $\mathcal \{H_0^i\}_{i=1}^n$ at $\mathcal X_s$ using \autoref{eq:initial_history_field}.\\
Obtain tensile strain energy, $\{\mathcal {H}_i^j\}_{i,j=1}^{n,r-1}$ at $\mathcal X_s$ from the high-fidelity data.\\
Compute the initial history function, $\{\mathbf H_0^j\}_{j=1}^n$ at $\{\bm y_i\}_{i=1}^p$ points.\\
Prepare the data for trunk net: $\left[\{\bm y_i\}_{i=1}^p,\{\mathbf H_0^j\}_{j=1}^n, \{\Delta \wb_k\}_{k=1}^r\right]$ as shown in \autoref{fig:unstacked_DeepONet}(b).\\
Construct the input data for branch net as defined in \autoref{eq:hist_window}. For each case, the tensor dimensions of the window is $(p,2m)$.\\
Initialize the V-DeepONet and the weights of the network using Xavier initialization technique.\\
Obtain the solution operators, $\mathcal G_{\bm \theta}$ and evaluate it at $\{\bm y_i\}_{i=1}^p$ using \autoref{eq:output_deeponets}.\\
Construct the hybrid loss-function using \autoref{eq:DeepOnet_loss}.\\
Minimize the hybrid loss and obtain the optimized parameters, $\bm \theta^*$.\\
For prediction, choose an initial configuration and obtain $\bm{\mathcal H}_0$ at $\mathcal X_s$ and $\mathbf H_0$ at $q$ random points sampled in the domain, and $t$ displacement steps.\\
\For {$i= 1,\ldots,t$}{
\eIf{$i==1$}{
Input to branch net = $\left[\bm{\mathcal H}_0\left(\mathcal X_s\right), \mathbf{0}\right]$.\\
Repeat this vector $q$ times.}
{Input to branch net = $\left[\bm{\mathcal H}_i, \bm{\mathcal H}_{i-1}\right]$.\\
Repeat this vector $q$ times.}
Input to the trunk net: $\left[\{\bm y_j\}_{j=1}^q,\mathbf H_0, \Delta \wb_i\right]$.\\
Predict the solutions, $u^*(\bm y_j), v^*(\bm y_j)$ and $\phi^*(\bm y_j)$.\\
Using the solutions compute the energy, $\bm{\mathcal H}_i$ required as input to the next step.\\
}\end{algorithm}
\FloatBarrier

\section{Flow in heterogeneous porous media}
\label{app:appendixA}

\begin{figure}[!htbp]
    \centering
    \subfigure[]{
    \includegraphics[trim=20 0 40 80,clip,width = \textwidth]{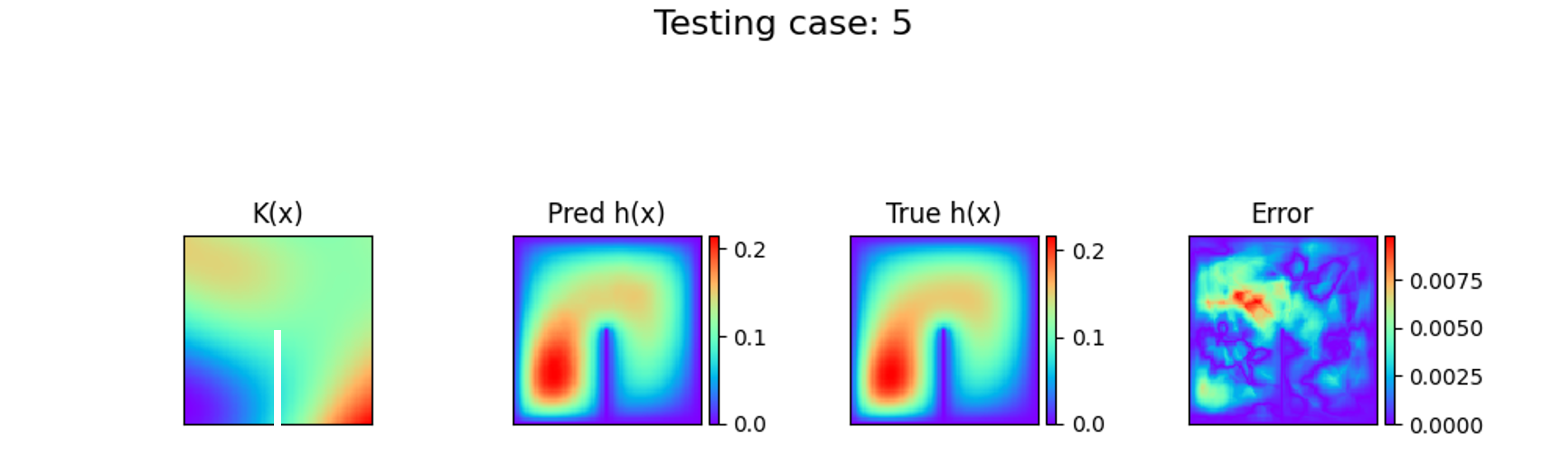}}
    \subfigure[]{
    \includegraphics[trim=20 0 40 80,clip,width = \textwidth]{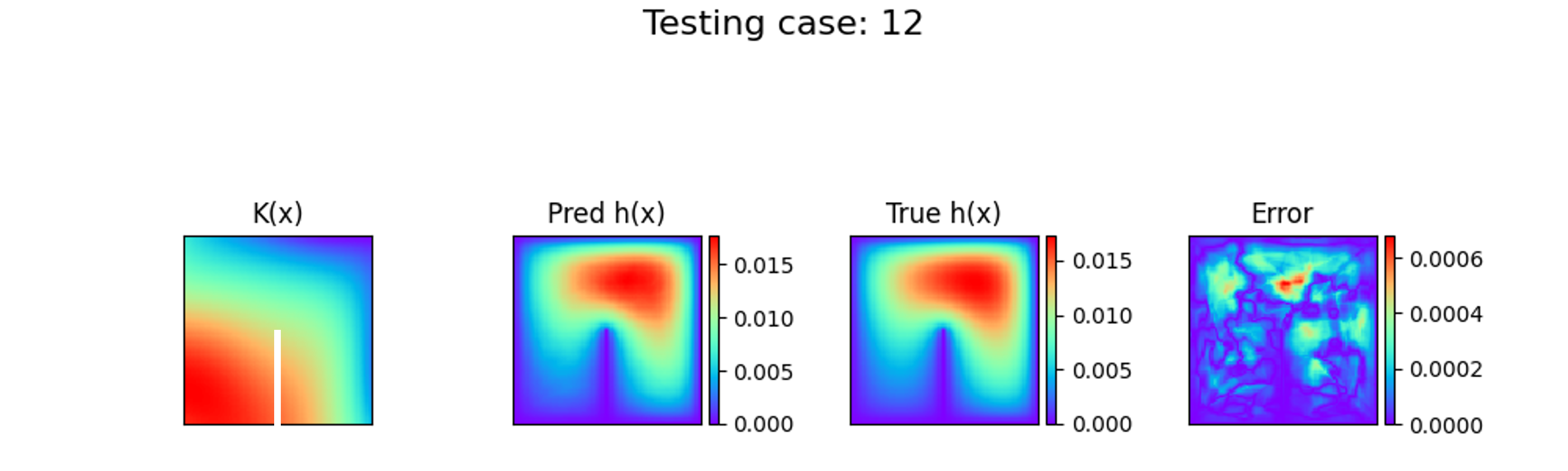}}
    \caption{Flow in heterogeneous porous media: The predicted $h(\bm x)$ for a given permeability, $K(\bm x)$ (plotted on log scale) ares shown for two samples. True $h(\bm x)$ represents the ground truth and is the simulated solution using the PDE toolbox. The difference between the predicted $h(\bm x)$ and the ground truth is shown in the error plot.}
    \label{fig:darcy_plots_test}
\end{figure}

We consider a two-dimensional flow through heterogeneous porous media, which is governed by the following equation:
\begin{equation}\label{eq:darcy_PDE}
\begin{split}
    -\nabla \cdot (K(\bm x)\nabla h(\bm x)) &= 1,\;\;\; \bm x = (x,y),\\
    \text{subjected to}\;\;\;h(\bm x) &= 0, \;\;\; \forall \;\; \bm x \in \partial \Omega,
\end{split}
\end{equation}
where $K(\bm x)$ is spatially varying hydraulic conductivity, and $h(\bm x)$ is the hydraulic head. In this example, we aim to learn the operator such that:
\begin{equation}
    \mathcal G_{\bm{\theta}}:K(\bm x) \rightarrow h(\bm x). 
\end{equation}
The setup is of a unit square plate with a discontinuity of $5\times10^{-3}$ mm. For generating multiple permeability field for training the V-DeepONet, we describe the conductivity field, $K(\bm x)$, as a stochastic process. In particular, we take 
$K(\bm x) = \exp(F(\bm x))$, with $F(\bm x)$ denoting a truncated Karhunen-Lo\`eve (KL) expansion for a certain Gaussian process, which is a finite-dimensional random variable. In our samples, the leading 100 terms in the KL expansion were kept for the Gaussian process with zero mean and the following kernel \cite{meng2021learning}:
\begin{equation}
\begin{split}
    &\mathcal K(\bm x, \bm x') = \mathcal K((x,y), (x',y')) = \exp \left[\frac{-(x-x')}{2l_1^2} + \frac{-(y-y')^2}{2l_2^2}\right],\\
    &\bm x,\bm x' \in [0,1]^2, l_1 = l_2 = 0.25. 
\end{split}
\end{equation}
In this example, the V-DeepONet is trained using the variational formulation, without any labelled input-output datasets. The optimization problem can be defined as:
\begin{equation}\label{eq:Darcy_energy}
\begin{split}
    &\text{Minimize:}\;\;\;\;\; \mathcal{E} = \Psi_h,\\
    &\text{subject to:}\;\;\;\; h(\bm{x}) = 0 \text{ on } \partial \Omega_{D},\\
    &\;\;\;\;\;\;\;\;\;\;\;\; 
\end{split}
\end{equation}
where
\begin{equation}\label{eq:energyterms_Darcy}
       \Psi_h = \frac{1}{2}\int_{\Omega}K(\bm x)|\nabla h(\bm x)|^2 \;\; d\Omega- \int_{\Omega}h(\bm x)\;\; d\Omega.
\end{equation}
We approximate the operator by a V-DeepONet architecture, where the branch and trunk networks are two separate 6-layer fully-connected neural networks with 32 neurons per hidden layer. In this example, $n = 200$ samples of permeability matrix is used to train. The solution operator is evaluated at $p = 10000$ randomly sampled points. 
The prediction of $h(\bm x)$ for two samples of $K(\bm x)$, using V-DeepONet is shown in \autoref{fig:darcy_plots_test}. The accuracy of V-DeepONet is verified by comparing the prediction of the network against the solution of \autoref{eq:darcy_PDE} obtained using the finite-element-based Partial Differential Equation Toolbox in Matlab using the same $K(\bm x)$. It is interesting to note that we have tried to solve the problem by minimizing the residual \cite{wang2021learning}. However, the residual based DeepONet is not able to approximate the solution of $h(\bm x)$ for a given $K(\bm x)$.
\FloatBarrier

\section{Additional results}
\label{app:appendixC}
\begin{figure}[!htbp]
    \centering
    \subfigure[$\Delta v_1 = 1.4\times 10^{-3}$ mm.]{
    \frame{\includegraphics[trim=40 30 0 20,clip,width = 0.45\textwidth]{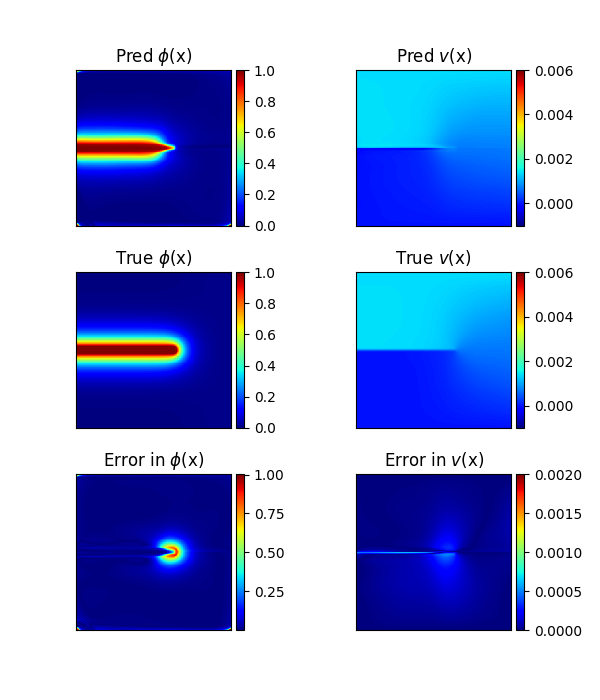}}}
    \subfigure[$\Delta v_4 = 5.0\times 10^{-3}$ mm.]{
    \frame{\includegraphics[trim=40 30 0 20,clip,width = 0.45\textwidth]{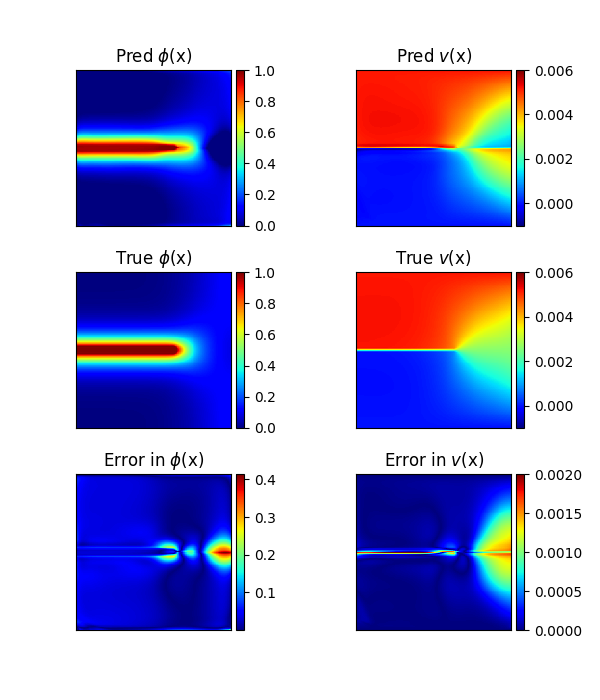}}}
    \subfigure[$\Delta v_5 = 5.6\times 10^{-3}$ mm.]{
    \frame{\includegraphics[trim=40 30 0 20,clip,width = 0.45\textwidth]{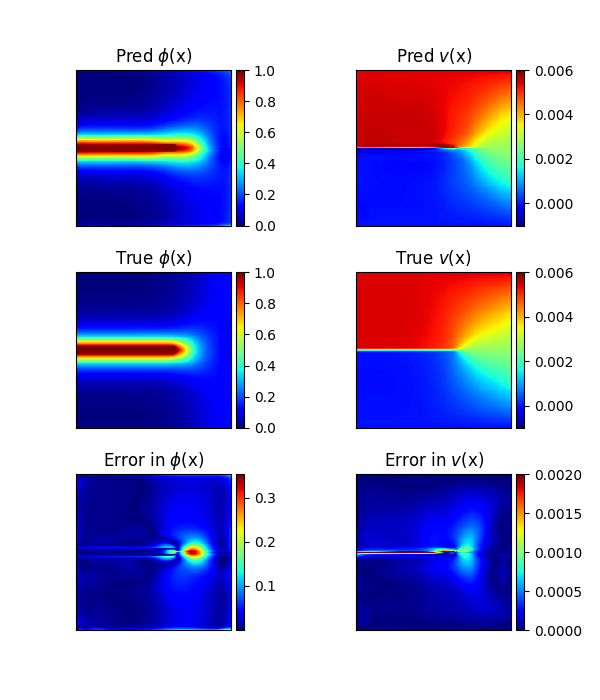}}}
    \subfigure[$\Delta v_6 = 5.8\times 10^{-3}$ mm.]{
    \frame{\includegraphics[trim=40 30 0 20,clip,width = 0.45\textwidth]{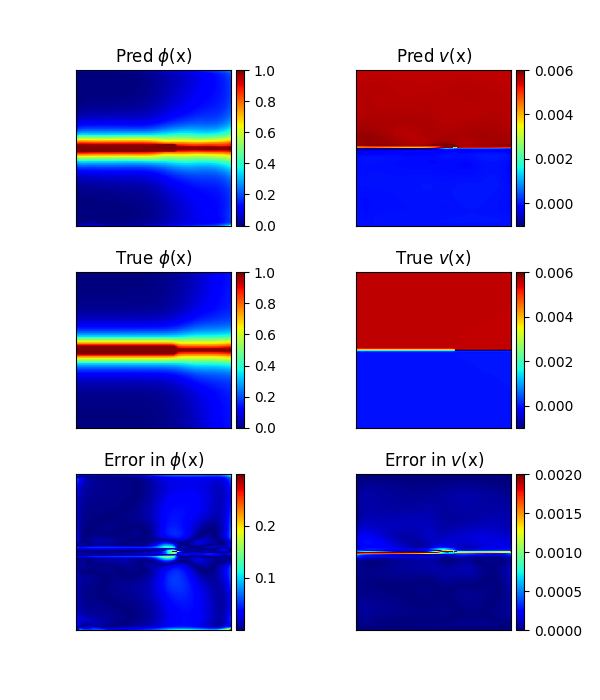}}}
    \caption{Tensile failure: V-DeepONet is trained with $6$ $l_c \in$ $[0.3,0.55]$ for $7$ displacement steps. In these plots, we present the predicted solution $l_c = 0.65$ mm (out-of-distribution) at four displacement steps, $\Delta v$, which are predicted sequentially. For each plot, the top row presents the predicted phase field and the displacement along \textit{y}-axis, respectively. The middle row shows the the ground truth obtained using the IGA simulations, while the last row shows the error between the predicted value and the ground truth.}
    \label{fig:tensile_testing_outofdist}
\end{figure}
\FloatBarrier

\begin{figure}
    \centering
    \includegraphics[width = 0.48\textwidth]{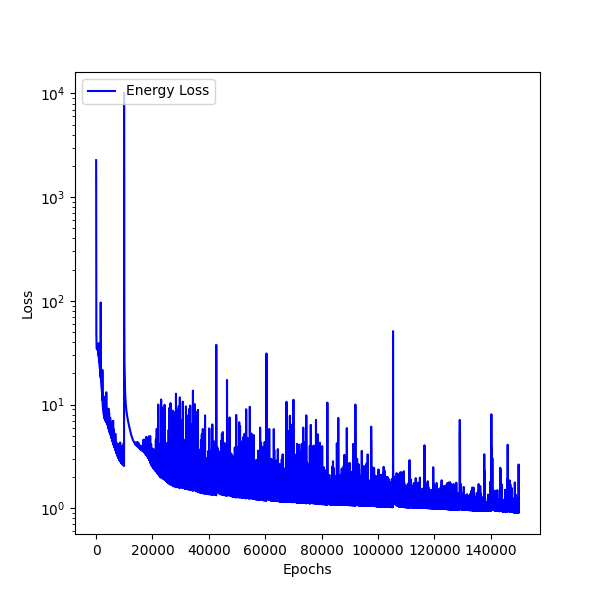}
    \caption{Shear failure: The training trajectory of V-DeepONet using $11$ training samples. The hybrid loss function in \autoref{eq:DeepOnet_loss} is minimized to obtain the optimized $\bm{\theta^*}$.}
    \label{fig:shear_X_error}
\end{figure}

\begin{figure}
    \centering
    \subfigure[$n=20$]{
    \frame{\includegraphics[trim=30 30 10 40,clip,width = 0.48\textwidth]{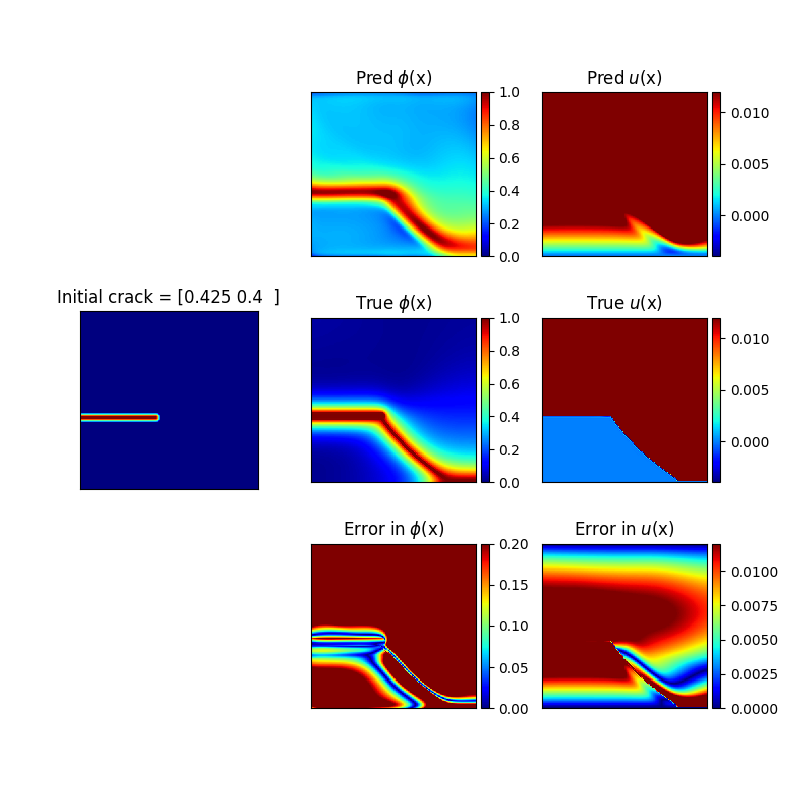}}}
    \subfigure[$n=40$]{
    \frame{\includegraphics[trim=30 30 10 40,clip,width = 0.48\textwidth]{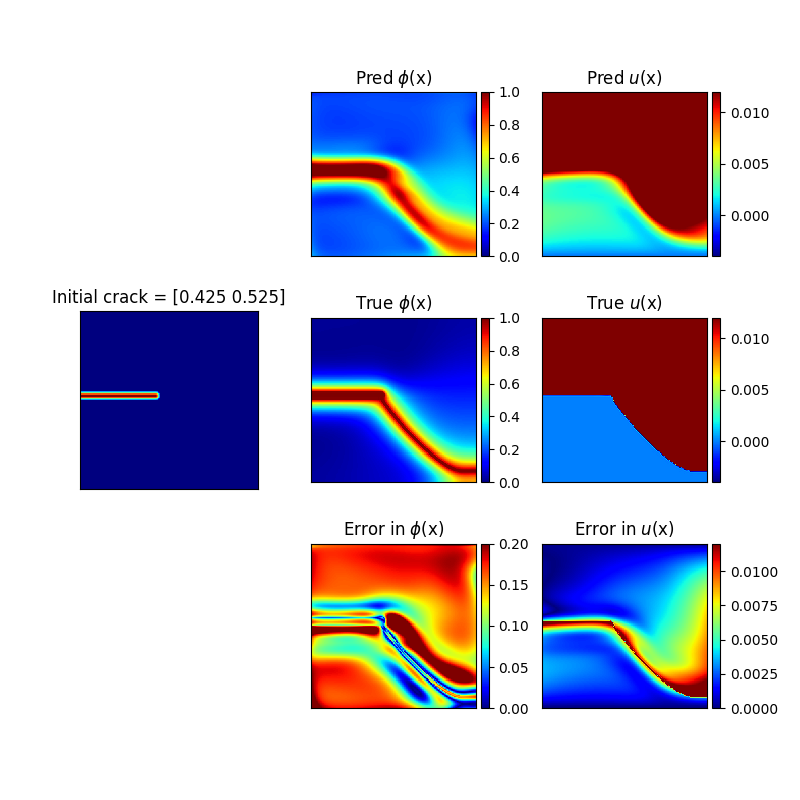}}}    
    \subfigure[$n=60$]{
    \frame{\includegraphics[trim=30 30 10 40,clip,width = 0.48\textwidth]{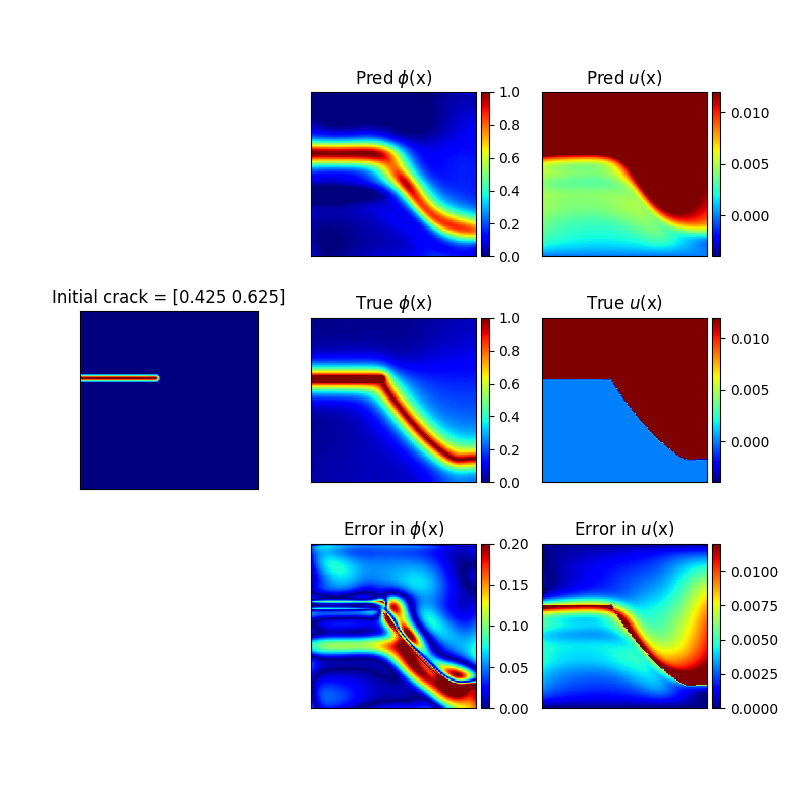}}}
    \subfigure[$n=85$]{
    \frame{\includegraphics[trim=30 30 10 40,clip,width = 0.48\textwidth]{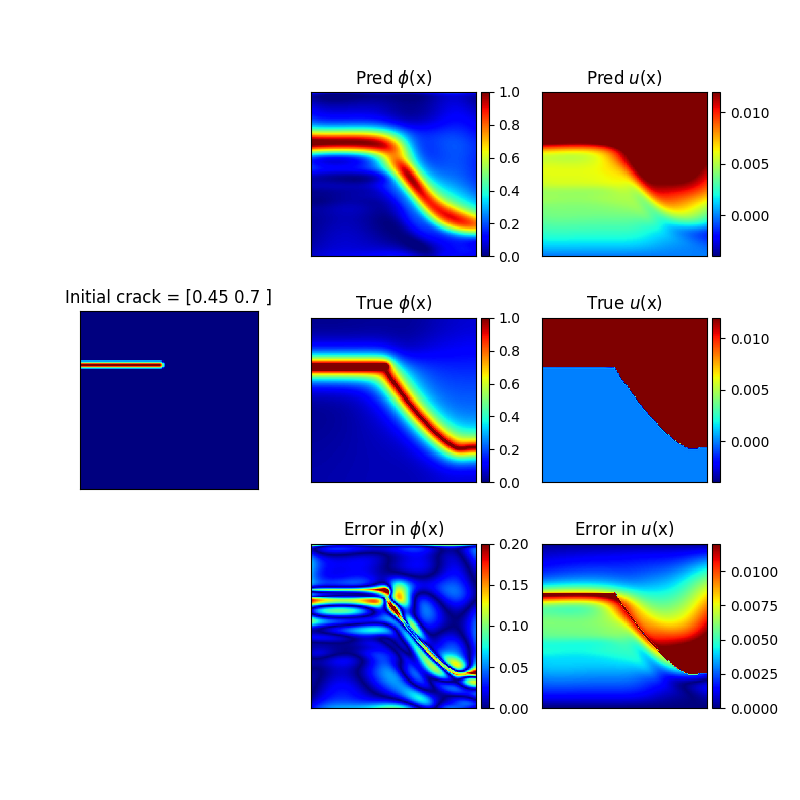}}}    
    \caption{Shear failure: V-DeepONet is trained with $20, 40$, $60$ and $85$ samples with crack tips located through out the domain. For each of the sample sizes, different crack tips beyond the training range have been considered for prediction. Using the trained model, we predict the final damage path and the displacement in $x$-axis for an out-of-distribution model for each of the cases.}
    \label{fig:shear_Ytest_OutofD}
\end{figure}
\end{appendices}

\bibliographystyle{elsarticle-num} 
\bibliography{cas-refs}

\end{document}